\pdfoutput=1

\documentclass[11pt]{article}

\usepackage{times}
\usepackage{latexsym}
\usepackage{graphicx}
\usepackage{comment}
\usepackage{lipsum}
\usepackage{footnote}
\usepackage{hyperref}
\usepackage{caption}
\usepackage{subcaption}
\usepackage{longtable}
\usepackage{pdflscape}
\usepackage{geometry}
\usepackage{booktabs}
\usepackage{wrapfig}
\usepackage{float}
\usepackage{pifont}
\usepackage{xcolor,colortbl}
\usepackage{multirow}
\usepackage[raggedrightboxes]{ragged2e}
\newcommand{\cmark}{\ding{51}}%
\newcommand{\xmark}{\ding{55}}%
\usepackage{tikz}
\usetikzlibrary{calc}
\usepackage{multirow}
\usepackage{longtable}
\usepackage{tabularx}
\usepackage{array}
\usepackage{adjustbox}
\usepackage{etaremune}
\usepackage{tcolorbox}
\usepackage{tikz,lipsum}
\usepackage{amsmath}
\usepackage{amssymb}

\usepackage[]{ACL2023}

\usepackage[T1]{fontenc}

\usepackage[utf8]{inputenc}

\newcommand{\act}[1]{\textcolor{blue}{#1}}

\usepackage[draft,textsize=footnotesize,textwidth=15mm]{todonotes}

\usepackage[T1]{fontenc}

\usepackage[utf8]{inputenc}

\usepackage{microtype}

\usepackage{inconsolata}

\title{
\includegraphics[width=0.75\textwidth]{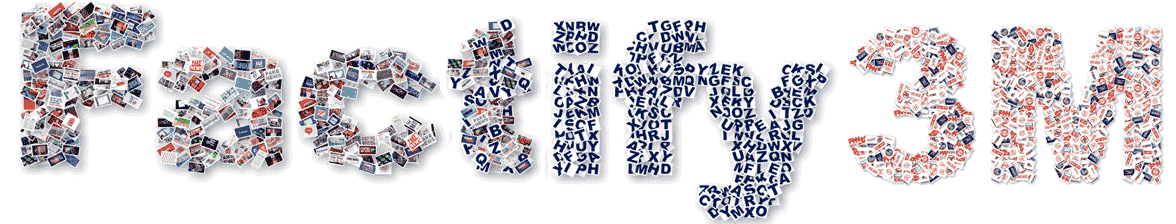}
\\
\vspace{-0.2cm}
A benchmark for multimodal fact verification with explainability through 5W Question-Answering}

\author{ 
Megha Chakraborty\textsuperscript{1} \ 
Khushbu Pahwa\textsuperscript{2} \
Anku Rani\textsuperscript{1}\ 
\bf Shreyas Chatterjee\textsuperscript{1}  \ 
\bf Dwip Dalal\textsuperscript{1} \ \\ 
\bf Harshit Dave\textsuperscript{1} \ 
\bf Ritvik G\textsuperscript{1}  \ 
\bf Preethi Gurumurthy\textsuperscript{1} \ 
\bf Adarsh Mahor\textsuperscript{1} \ 
\bf Samahriti Mukherjee\textsuperscript{1} \ \\
\bf Aditya Pakala\textsuperscript{1} \ 
\bf Ishan Paul\textsuperscript{1} \ 
\bf Janvita Reddy\textsuperscript{1} \
\bf Arghya Sarkar\textsuperscript{1} \
\bf Kinjal Sensharma\textsuperscript{1} \\\ 
\textbf{Aman Chadha}\textsuperscript{3,4\dag} \
\bf Amit P. Sheth\textsuperscript{1} \ 
\bf Amitava Das\textsuperscript{1} \ \\
\textsuperscript{1} \small University of South Carolina, USA \
\textsuperscript{2} \small UCLA, USA \
\textsuperscript{3} \small Amazon AI, USA \
\textsuperscript{4} \small Stanford University, USA\\
\small \tt meghac@email.sc.edu \
\small \tt amitava@mailbox.sc.edu
}

\begin{document}
\maketitle

\renewcommand{\thefootnote}{\fnsymbol{footnote}}
\footnotetext[2]{Work does not relate to position at Amazon.}
\renewcommand*{\thefootnote}{\arabic{footnote}}
\setcounter{footnote}{0}

\begin{abstract}

 Combating disinformation is one of the burning societal crises - about 67\% of the American population believes that disinformation produces a lot of uncertainty, and 10\% of them knowingly propagate disinformation. Evidence shows that disinformation can manipulate democratic processes and public opinion, causing disruption in the share market, panic and anxiety in society, and even death during crises. Therefore, disinformation should be identified promptly and, if possible, mitigated. With approximately 3.2 billion images and 720,000 hours of video shared online daily on social media platforms, scalable detection of multimodal disinformation requires efficient fact verification. Despite progress in automatic text-based fact verification (e.g., FEVER, LIAR), the research community lacks substantial effort in multimodal fact verification. To address this gap, we introduce FACTIFY 3M, a dataset of 3 million samples that pushes the boundaries of the domain of fact verification via a multimodal fake news dataset, in addition to offering explainability through the concept of 5W question-answering. Salient features of the dataset include: \textit{(i) textual claims, (ii) ChatGPT-generated paraphrased claims, (iii) associated images, (iv) stable diffusion-generated additional images (i.e., visual paraphrases), (v) pixel-level image heatmap to foster image-text explainability of the claim, (vi) 5W QA pairs, and (vii) adversarial fake news stories.}
\end{abstract}

\section{ FACTIFY 3M - an illustration}
\label{sec:intro}
We introduce FACTIFY 3M (\textit{3 million}), the largest dataset and benchmark for multimodal fact verification.

\begin{figure}[!tbh]
\centering
\includegraphics[width=0.7\columnwidth]{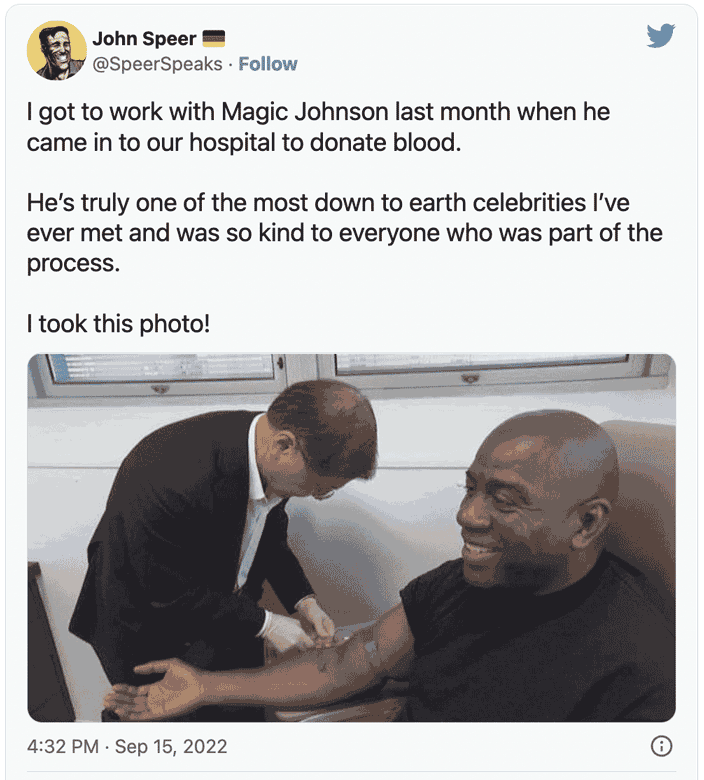}
\caption{A tweet referring to the sports personality and known AIDS victim Magic Johnson with a photo that was taken a decade before COVID and, moreover, is not real evidence of blood donation.}
\label{fig:magic_johnson}
\end{figure}

\begin{table*}[!h]
\centering
\resizebox{0.99\textwidth}{!}{%
\begin{tabular}{@{}llccccccccc@{}}
\toprule
\multicolumn{11}{c}{\cellcolor[HTML]{CBCEFB}\textbf{PromptFake3M at a glance}} \\ \midrule
\multicolumn{2}{c}{\textbf{Entailment classes}} &
  \textbf{Textual Support} &
  \textbf{Visual/Image Support} &
  \textbf{No. of claims} &
  \textbf{\begin{tabular}[c]{@{}c@{}}No. of paraphrased \\ claims\end{tabular}} &
  \textbf{No. of images} &
  \textbf{\begin{tabular}[c]{@{}c@{}}No. of stable diffusion \\ generated images\end{tabular}} &
  \textbf{5WQA pairs} &
  \textbf{\begin{tabular}[c]{@{}c@{}}No. of evidence \\ documents\end{tabular}} &
  \textbf{\begin{tabular}[c]{@{}c@{}}Adversarial\\  OPT-generated news story\end{tabular}} \\
  \hline
 &
  \cellcolor[HTML]{009901}\textbf{\textcolor{white}{Support\_Multimodal}} &
  \begin{tabular}[c]{@{}c@{}}Texts are supporting\\ each other \\ $\sim$similar news\end{tabular} &
  \begin{tabular}[c]{@{}c@{}}Images are \\ supporting each other\end{tabular} 
    & 232,000
    & 882,000
    & 232,000
    & 927,000
    & 858,400
    & 232,000
    & \xmark \\
  \cline{2-11} 

\multirow{-3}{*}{\textbf{\rotatebox{90}{\textcolor{green}{{\textbf{Support}}}}}} &
  \cellcolor[HTML]{009901}\textbf{\textcolor{white}{Support\_Text}} &
  \begin{tabular}[c]{@{}c@{}}Texts are supporting\\ each other\\ $\sim$similar news\end{tabular} &
  \begin{tabular}[c]{@{}c@{}}Images are neither \\ supporting nor refuting\end{tabular} 
   & 174,000
   & 609,000
   & 169,000
   & 661,000
   & 852,600
   & 174,000
   & \xmark \\
  \cline{2-11}

 &
  \cellcolor[HTML]{F8A102}\textbf{\textcolor{white}{Insufficient\_Multimodal}} &
  \begin{tabular}[c]{@{}c@{}}Texts are neither \\ supported nor refuted\\ $\sim$may have common words\end{tabular} &
  \begin{tabular}[c]{@{}c@{}}Images are \\ supporting each other\end{tabular} 
   & 99,000
   & 366,000
   & 99,000
   & 347,000
   & 375,000
   & 99,000
   & \xmark \\
  \cline{2-11} 

\multirow{-3}{*}{\textbf{\rotatebox{90}{\textcolor{orange}{\textbf{Neutral}}}}} &
  \cellcolor[HTML]{F8A102}\textbf{\textcolor{white}{Insufficient\_Text}} &
  \begin{tabular}[c]{@{}c@{}}Texts are neither \\ supported nor refuted\\ $\sim$may have common words\end{tabular} &
  \begin{tabular}[c]{@{}c@{}}Images are neither \\ supporting nor refuting\end{tabular} 
   & 126,000
   & 525,000
   & 123,000
   & 466,000
   & 441,000
   & 126,000
   & \xmark \\
  \cline{2-11} 
 \textbf{\rotatebox{90}{\textcolor{red}{\textbf{Fake}}}} &
  \cellcolor[HTML]{E60A0A}{\textbf{\textcolor{white}{Refute}}}  &
  \begin{tabular}[c]{@{}c@{}}Fake claim \\\end{tabular}  &
  \begin{tabular}[c]{@{}c@{}}Fake image support \\\end{tabular}  
   & 316,000
   & 1,193,000
   & 309,000
   & 916,400
   & 1,327,000
   & 316,000
   & \cmark 135,000\\
   \hline
   \hline
\multicolumn{2}{l}{Total} &
  \multicolumn{1}{c}{} &
  \multicolumn{1}{c}{} &
  \multicolumn{1}{c}{947,000} &
  \multicolumn{1}{c}{3,575,000} &
  \multicolumn{1}{c}{932,000} &
  \multicolumn{1}{c}{3,317,400} &
  \multicolumn{1}{c}{3,954,000} &
  \multicolumn{1}{c}{947,000} &
  \multicolumn{1}{c}{135,000} \\ \bottomrule
\end{tabular}%
}
\caption{A top-level view of FACTIFY 3M: (i) classes and their respective textual/visual support specifics, (ii) number of claims, paraphrased claims, associated images, generated images, 5W pairs, evidence documents, and adversarial stories.}
\label{tab:glance}
\end{table*}

\begin{figure*}[!ht]
\centering
\resizebox{0.97\textwidth}{!}{%
\begin{tabular}{@{}lllll@{}}
\multicolumn{5}{c}{\cellcolor[HTML]{68CBD0}{
\begin{huge}
\textbf{5W QA based Explainability}
\end{huge}
}}                                       \\ \midrule
\begin{LARGE}\textbf{Who claims}\end{LARGE}     & \begin{LARGE}\textbf{What claims}\end{LARGE}    & \begin{LARGE}\textbf{When claims}\end{LARGE}    & \begin{LARGE}\textbf{Where claims}\end{LARGE}   & \begin{LARGE}\textbf{Why claims}\end{LARGE}    
\\
\hline
\begin{tabular}[c]{@{}l@{}}
\parbox{5cm}{
\begin{Large}
\begin{itemize}
\item \textbf{Q1}: \textbf{\textit{Who went to the hospital?}}\\ \underline{Ans}: Magic Johnson
\item \textbf{Q2}: \textbf{\textit{Who worked with whom?}}\\ \underline{Ans}: the author with Magic Johnson
\item \textbf{Q3}: \textbf{\textit{Who took the photo?}}\\ \underline{Ans}: the author
\end{itemize}
\end{Large}
}
\end{tabular} &
  \begin{tabular}[c]{@{}l@{}}
  \parbox{5cm}{
\begin{Large}
  \begin{itemize}
  \item \textbf{Q1}: \textbf{\textit{What did Magic Johnson do at the hospital?}}\\
  \underline{Ans:} donated blood
  \item \textbf{Q2}: \textbf{\textit{What process Magic Johnson was part of?}}\\
  \underline{Ans}: blood donation
  \end{itemize}
\end{Large}
  }
  \end{tabular} &
  \begin{tabular}[c]{@{}l@{}}
  \parbox{5cm}{
\begin{Large}
  \begin{itemize}
  \item \textbf{Q1}: \textbf{\textit{When did Magic Johnson visit the hospital?}}\\ \underline{Ans}: last month \\ time of the post = Sept - 1 month from August
  \end{itemize}
\end{Large}
  }
  \end{tabular} &
  \begin{tabular}[c]{@{}l@{}}
  \parbox{5cm}{
\begin{Large}
  \begin{itemize}
  \item \textbf{Q1}: \textbf{\textit{Where did Magic Johnson pay visit to?}}\\ \underline{Ans}: hospital.
  \end{itemize}
\end{Large}
  }
  \end{tabular} &
  \begin{tabular}[c]{@{}l@{}}
  \parbox{5cm}{
\begin{Large}
  \begin{itemize}
  \item \textbf{Q1}: \textbf{\textit{Why did Magic Johnson visit hospital?}}\\ \underline{Ans}: to donate blood.
  \end{itemize}
\end{Large}
  }
  \end{tabular}\\
  \hline
\includegraphics[width=0.08\textwidth]{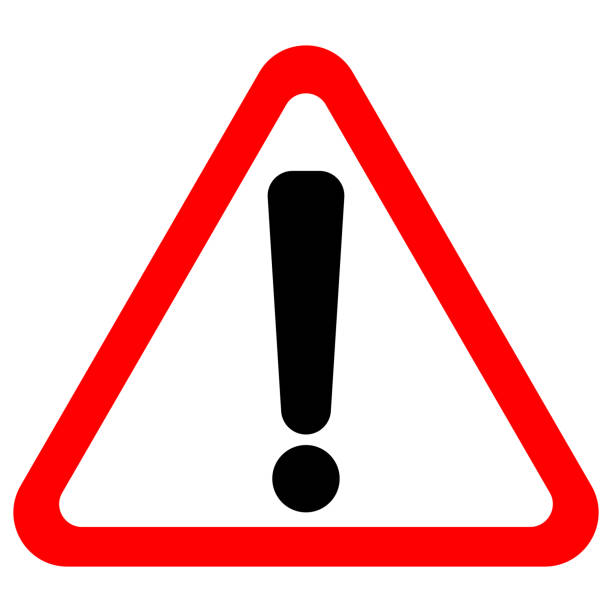} 
\begin{LARGE}\textbf{\textcolor{orange}{caution}}\end{LARGE}
& 
\includegraphics[width=0.06\textwidth]{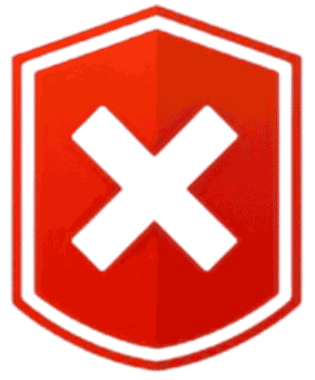} 
\begin{LARGE}\textbf{\textcolor{red}{verified false}}\end{LARGE}
& 
\includegraphics[width=0.08\textwidth]{img/caution.jpeg} 
\begin{LARGE}\textbf{\textcolor{orange}{caution}}\end{LARGE}
& 
\includegraphics[width=0.04\textwidth]{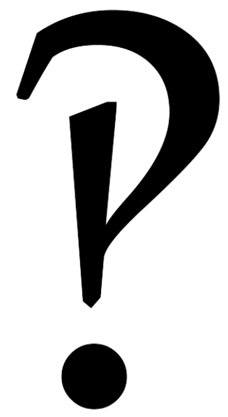} 
\begin{LARGE}\textbf{not verifiable}\end{LARGE}
& 
\includegraphics[width=0.06\textwidth]{img/false.png} 
\begin{LARGE}\textbf{\textcolor{red}{verified false}}\end{LARGE}
\\
\hline
\multicolumn{5}{c}{\cellcolor[HTML]{C0C0C0}{
\begin{huge}
\textbf{Evidence}
\end{huge}
}
}  
\\
\hline
\begin{tabular}[c]{@{}l@{}}
\parbox{5cm}{
\begin{Large}
\begin{itemize}
\item \textbf{news1} - \textcolor{blue}{url1} - Magic Johnson visits hundreds of kids at SC hospital, on \underline{Dec 10, 2019} 
\end{itemize}
\begin{itemize}
\item No information is available that the author worked with Magic Johnson
\item No information is available about who took this photo.
\end{itemize}
\end{Large}
}
\end{tabular} &
  \begin{tabular}[c]{@{}l@{}}
  \parbox{5cm}{
\begin{Large}
  \begin{itemize}
  \item \textbf{news1} - \textcolor{blue}{url1} - Magic Johnson Shuts Down '\textit{False Story}' He Donated Blood
  \item \textbf{news 2} - \textcolor{blue}{url2} - Magic Johnson \textbf{\underline{didn't donate blood}} to help fight COVID-19
  \end{itemize}
  
  related story - 
  \begin{itemize}
  \item \textbf{news 3} - \textcolor{blue}{url3} - Magic Johnson opens up on living with HIV 30 years
  \end{itemize}
\end{Large}
  }
  \end{tabular} &
  \begin{tabular}[c]{@{}l@{}}
  \parbox{5cm}{
\begin{Large}
  \begin{itemize}
  \item \textbf{news1} - \textcolor{blue}{url1} - Magic Johnson visits hundreds of kids at SC hospital, on \\ \underline{Dec 10, 2019} 
  \item \textbf{news2} - \textcolor{blue}{url2} - \\ Magic Johnson at Howard University Hospital, on \underline{Feb 7, 2013}
\end{itemize}
\end{Large}
  }
  \end{tabular} &
  \begin{tabular}[c]{@{}l@{}}
  \parbox{5cm}{
\begin{Large}
  \begin{itemize}
\item \textbf{news1} - \textcolor{blue}{url1} - Magic Johnson visits hundreds of kids at SC hospital, on \\ \underline{Dec 10, 2019} 
\item \textbf{news2} - \textcolor{blue}{url2} - \\ Magic Johnson at Howard University Hospital, on \underline{Feb 7, 2013}
  \end{itemize}
\end{Large}
  }
  \end{tabular} &
  \begin{tabular}[c]{@{}l@{}}
  \parbox{5cm}{
\begin{Large}
  \begin{itemize}
  \item \textbf{news1} - \textcolor{blue}{url1} - Magic Johnson Shuts Down '\textit{False Story}' He Donated Blood
  \item \textbf{news 2} - \textcolor{blue}{url2} - Magic Johnson \textbf{\underline{didn't donate blood}} to help fight COVID-19
  \end{itemize}
\end{Large}
  }\end{tabular} \\ \hline
\end{tabular}%
}
\caption{An illustration of the proposed 5W QA-based explainable fact verification system. This example illustrates the false claim shown in fig. ~\ref{fig:magic_johnson}. A typical semantic role labeling (SRL) system processes a sentence and identifies verb-specific semantic roles. Therefore, for the specified example, we have 3 sentences: sentence 1 has two main verbs \textit{work} and \textit{come}, sentence 2 has one verb \textit{meet}, and sentence 3 has one verb \textit{take}. For each verb, a 5W QA pair will be automatically generated (4 $\times$ 5 = 20 sets of QA pairs in total for this example). Furthermore, all those 20 5W aspects will be fact-checked. If some aspects end up having \textit{neutral} entailment verdict, possible relevant documents with associated URLs will be listed for the end user to read further and assess. In addition, a reverse image search result will be shown to aid human fact-checkers further.}
\label{fig:5WQA_example_1}

\end{figure*}

\begin{figure}
    \centering
    \resizebox{0.8\columnwidth}{!}{%
{\small
\fbox{%
    \parbox{\columnwidth}{%
    \textcolor{blue}{Sports star Magic Johnson came to the hospital last month to donate blood to support the COVID-19 crisis.}
    \\
    \textbf{Prphr 1:} Last month, Magic Johnson, a famous athlete, visited a hospital to give blood to help with the COVID-19 pandemic.
     \\
    \textbf{Prphr 2:} Magic Johnson, a well-known sports figure, went to a hospital last month to donate blood to assist with the COVID-19 crisis. 
     \\
    \textbf{Prphr 3:} Magic Johnson, a sports celebrity, recently visited a hospital and donated blood to contribute to the COVID-19 crisis.
     \\
    \textbf{Prphr 4:} Recently, Magic Johnson, a well-known sports star, went to a hospital and gave blood to support the COVID-19 crisis.   
     \\
    \textbf{Prphr 5:} To support the COVID-19 crisis, Magic Johnson, a sports celebrity, came to a hospital last month to donate blood.  
     \\
    \textbf{Prphr 6:} In an effort to support the COVID-19 crisis, Magic Johnson, a well-known sports figure, visited a hospital last month to donate blood. 
    }
    }%
    }
}
\caption{Claims paraphrased using ChatGPT \cite{chatgpt} to foster their textual diversity.}
\label{fig:chatgpt-image}
\end{figure}

Consider the example in fig. \ref{fig:magic_johnson}. A widely distributed image of the sports legend Magic Johnson with an IV line in his arm was accompanied by the claim that he was donating blood, implicitly during the COVID-19 pandemic. If true, this is troubling because Magic Johnson is a well-known victim of AIDS and is prohibited from donating blood. The picture predated the COVID-19 epidemic by a decade and is related to his treatment for AIDS.  

\textbf{Textual claim:} The text associated with the claim in fig. \ref{fig:magic_johnson} purports that the author of this tweet took this photo and assisted Magic Johnson in donating blood. The further implicit declaration is that he is a medical worker and possibly works for a hospital that Magic Johnson visited for blood donation.

\begin{figure}[!tbh]
\centering
\begin{subfigure}[b]{0.46\textwidth}
\centering
\includegraphics[width=0.8\textwidth]{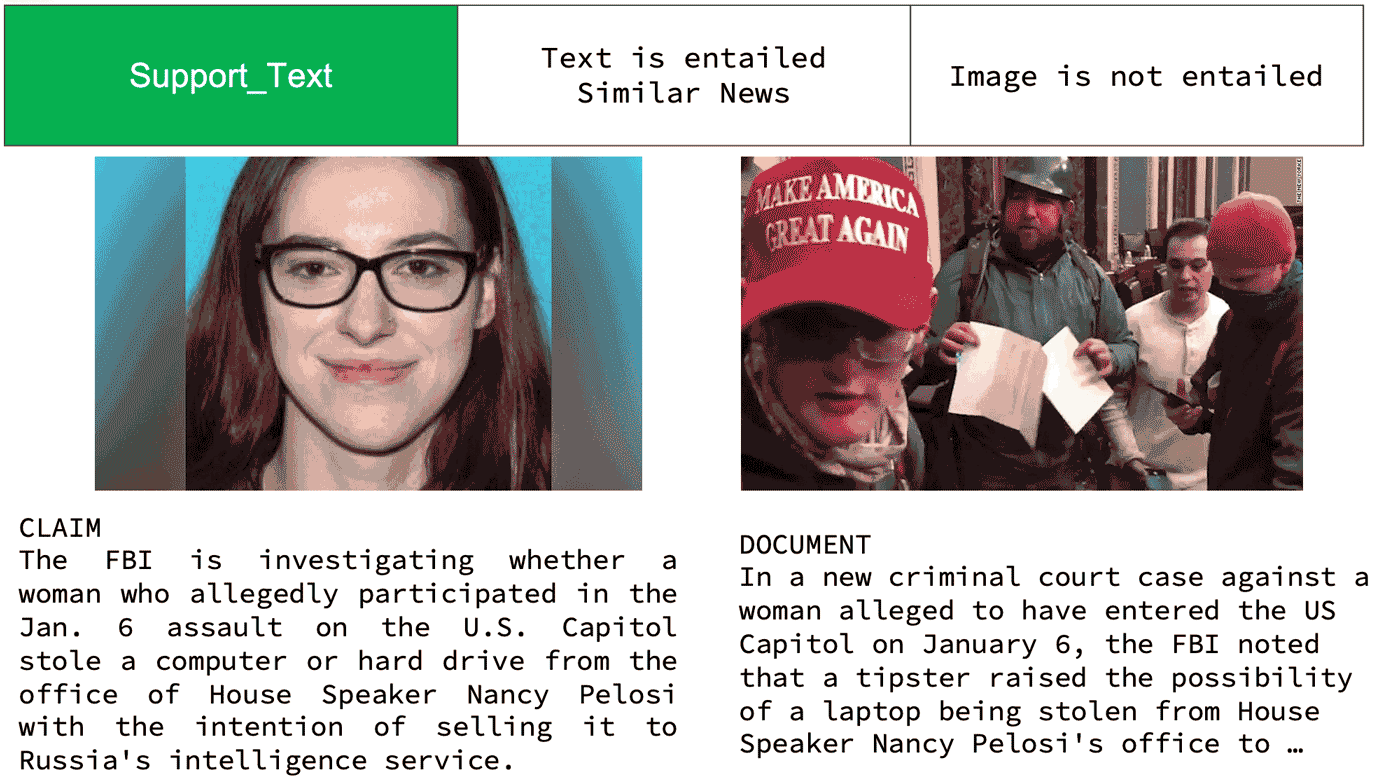}
\caption{Another example of covering the same news event by two news media houses. Here the same alleged lady is visible in both images, but altogether two images are different, and the text is paraphrased differently.}
\label{fig:claim_venice}
\end{subfigure}
\\
\begin{subfigure}[!tbh]{0.46\textwidth}
\centering
\includegraphics[width=0.8\textwidth]{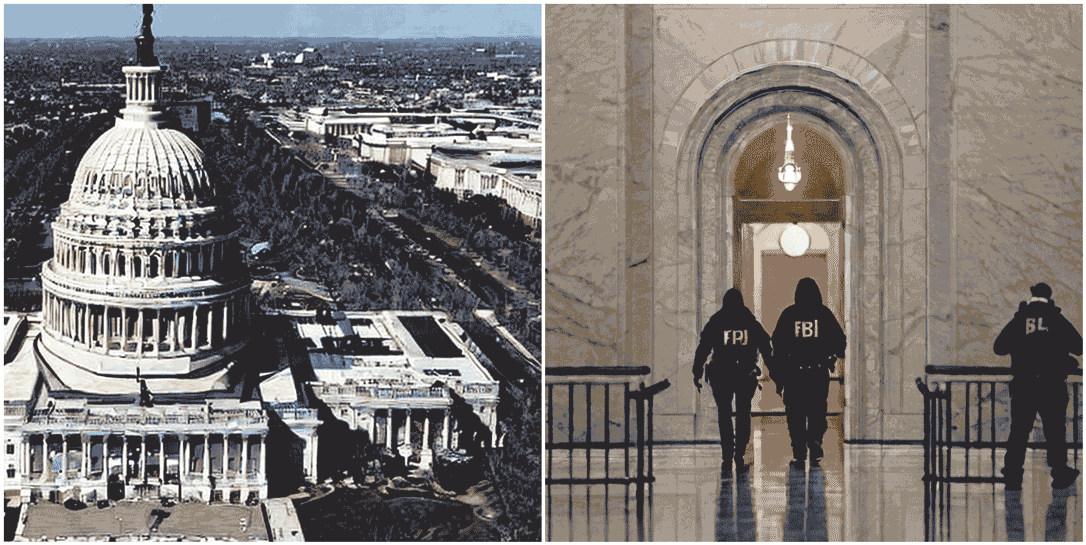}
\caption{Stable Diffusion output for the above claim.}
\label{fig:sd_venice}
\end{subfigure}%
\\
\begin{subfigure}[!tbh]{0.46\textwidth}
\centering
\includegraphics[width=0.8\textwidth]{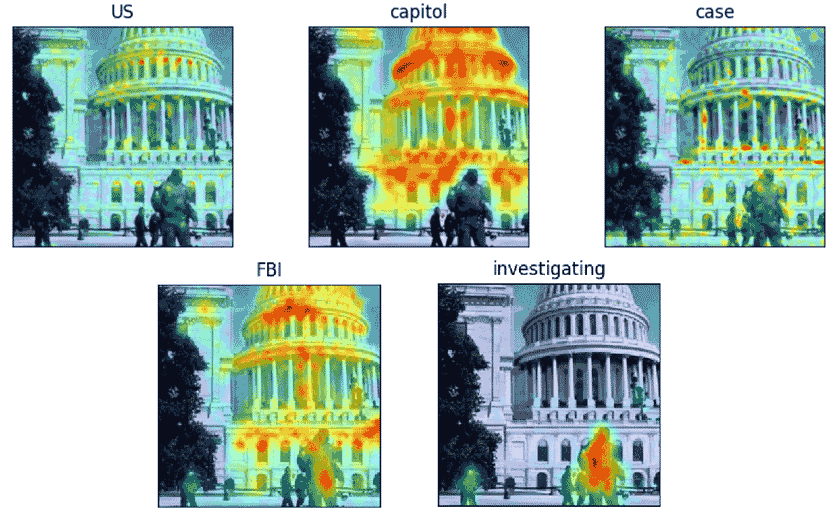}
\caption{DAAM \cite{tang2022daam} explanation for the above claim.}
\label{fig:daam_venice}
\end{subfigure}
\caption{An example from PromptFake 3M dataset. Multimodal fact verification is a challenging endeavor considering the intricacies of real-life data, where the entailment of images requires understanding the nuances of day-to-day life situations. As such, multimodal entailment is an under-research paradigm with limited progress; current SoTA systems lack the finesse needed to handle the complexity portrayed in these previous examples adeptly.}
\label{fig:examples}
\end{figure}

\textbf{ChatGPT-paraphrased claims:} To emulate the textual diversity in how news publishing houses report day-to-day happenings, it is important to produce data representative of such variety. Variety in text is embodied through a difference in narrative styles, choice of words, ways of presenting factual information, etc. For e.g., fig. ~\ref{fig:claim_venice} shows a claim and document that address the same topic but differ in their textual prose. To mimic this aspect, we adopted ChatGPT as a paraphraser and generated claims. Fig. \ref{fig:chatgpt-image} shows an example of a claim paraphrased using ChatGPT \cite{chatgpt}.

\textbf{Associated images:} The image included as part of the claim (refer fig. \ref{fig:magic_johnson} for the image embedded in the tweet and fig. ~\ref{fig:claim_venice} for images included as part of the claim) improves its trustworthiness perception since humans tend to believe visual input much more than mere text prose. Moreover, the text and image components together provide a holistic claim assertion, similar to how news articles convey world updates.

\textbf{Stable Diffusion-generated additional images aka visual paraphrases:} Referring to fig. ~\ref{fig:claim_venice}, the diversity of images associated with the claim and document are apparent. Specifically, the image associated with the claim is that of a frontal face, while that associated with the document, while still the same person, is one of a not-so-visible face but includes multiple other entities. Such diversity is commonly seen in the wild when different news media houses cover the same news. As such, we try to emulate this aspect in our work by harnessing the power of the latest in text-to-image generation. We generate additional images using Stable Diffusion \cite{rombach2021highresolution}. Fig. ~\ref{fig:sd_venice} shows the diversity of generated images in terms of the camera angle, subject, entities, etc., which in turn offers enhanced visual diversity for a multimodal claim.

\textbf{Pixel-level image heatmap:} To clearly delineate and hence explain which aspect of the image is being referred to in the various components of the text caption, we generate pixel-level attribution heatmaps to foster explainability. For e.g., referring to fig. ~\ref{fig:daam_venice}, the heatmap highlights gothic-architecture buildings for the word \textit{capitol}, which the building is known for, and highlights the human figure for the term \textit{investigating}. A dataset of this kind would be very helpful in designing explainable multimodal fact verification and possibly visual question-answer-based fact verification.

\begin{figure}
    \centering
\resizebox{0.8\columnwidth}{!}{%
{\small
\fbox{%
    \parbox{\columnwidth}{%
        Sports star Magic Johnson came to the hospital last month to donate blood to support the COVID-19 crisis. \textcolor{blue}{He shared his experience on social media, expressing his appreciation to the medical staff at the hospital and encouraging others to donate blood. He said, “I just donated blood at the hospital to help those affected by the coronavirus pandemic. I want to thank all the doctors, nurses, and medical staff for their hard work and dedication. I urge everyone to step up and donate blood if they are able. Every donation makes a difference.”}
    }%
    }
    }
}
\caption{An example of OPT \cite{zhang2022opt} generated fake news that confirms the Magic Johnson blood donation incident.}
\label{fig:opt-image}
\end{figure}

\textbf{5WQA:} The process of fact verification is inherently intricate, with several questions representing the components within the underlying claim that need answers to reach a verdict on the veracity of the claim. Referring to the example in Fig. \ref{fig:magic_johnson}, such questions may include: \textit{(a) who donated blood? (b) when did he donate blood? (c) can Magic Johnson donate blood? (d) what can go wrong if this claim is false?} Manual fact-checking can be labor-intensive, consuming several hours or days \cite{10.1145/2806416.2806652, Adair2017ProgressT}.

Contemporary automatic fact-checking systems focus on estimating truthfulness using numerical scores, which are not human-interpretable. Others extract explicit mentions of the candidate's facts in the text as evidence for the candidate's facts, which can be hard to spot directly. Only two recent works \cite{9747214, kwiatkowski-etal-2019-natural} propose question answering as a proxy to fact verification explanation, breaking down automated fact-checking into several steps and providing a more detailed analysis of the decision-making processes. Question-answering-based fact explainability is indeed a very promising direction. However, open-ended QA for a fact can be hard to summarize. Therefore, we refine the QA-based explanation using the 5W framework (\textit{who, what, when, where, and why}). Journalists follow an established practice for fact-checking, verifying the so-called 5Ws \cite{10.2307/1023893}, \cite{stofer2009sports}, \cite{silverman}, \cite{su2019study}, \cite{smarts_2017}, \cite{article_2023}. This directs verification search and, moreover, identifies missing content in the claim that bears on its validity. One consequence of journalistic practice is that claim rejection is not a matter of degree (as conveyed by popular representations such as a number of Pinocchios or crows, or true, false, half true, half false, pants on fire), but the rather specific, substantive explanation that recipients can themselves evaluate \cite{dobbs2012rise}. Please refer to fig. ~\ref{fig:5WQA_example_1} to look at the possible 5W QA questionnaire for the claim in fig. \ref{fig:magic_johnson}.

\textbf{Adversarial fake news:} Fact verification systems are only as good as the evidence they can reference while verifying a claim's authenticity. Over the past decade, with social media having mushroomed into the masses' numero-uno choice of obtaining world news, fake news articles can be one of the biggest bias-inducers to a person's outlook towards the world. To this end, using the SoTA language model, we generate adversarial news stories to offer a new benchmark that future researchers can utilize to certify the performance of their fact verification systems against adversarial news.

Programmatic detection of AI-generated writing (where an AI is the sole author behind the article) and its more challenging cousin -- AI-assisted writing (where the authorship of the article is split between an AI and a human-in-the-loop) -- has been an area of recent focus. While detecting machine-generated text from server-side models (for instance, GPT-3 \cite{brown2020language}, which is primarily utilized through an API, uses techniques like watermarking \cite{tc-gpt}) is still a topic of investigation, being able to do so for the plethora of open-source LLMs available online is a herculean task. Our adversarial dataset will offer a testbed so that such detection advances can be measured against with the ultimate goal of curbing the proliferation of AI-generated fake news.

\section{Related works - data sources and compilation}
\label{sec:related}

Automatic fact verification has received significant attention in recent times. Several datasets are available for text-based fact verification, e.g., FEVER \cite{thorne2018fever}, Snopes \cite{vo2020facts}, PolitiFact \cite{vo2020facts}, FavIQ \cite{kwiatkowski-etal-2019-natural}, HoVer \cite{jiang2020hover}, X-Fact \cite{gupta2021xfact}, CREAK \cite{onoe2021creak}, FEVEROUS \cite{Aly21Feverous}, etc.

Multimodal fact verification has recently started gaining momentum. DEFACTIFY workshop series at AAAI 2022 \cite{factify1} and 2023 \cite{factify2} has released FACTIFY 1.0 \cite{mishra2022factify} and 2.0 \cite{surya2023factify2} with 50K annotated data each year, which we have embedded as well as part of FACTIFY 3M.

Fact verification datasets are mainly classified into three major categories: \textit{(i) support, (ii) neutral, and (iii) refute}. While it is relatively easier to collect data for \textit{support} and \textit{neutral} categories, collecting large-scale \textit{refute} category aka fake news/ claims is relatively challenging. FEVER \cite{thorne2018fever} - proposed an alternative via manual imaginary claim generation - is complex, strives with scalability issues and this process may generate something unrealistic. Therefore, we decided to merge available datasets, at least for the \textit{refute} category. It is wise to have all these datasets into one, and further, we have generated associated images using stable diffusion. While selecting datasets, we only chose datasets with evidence claim documents as we are interested in 5W QA-based explanation. Other datasets only with fake claims were discarded for this reason. Furthermore, we use OPT \cite{zhang2022opt} to generate adversarial fake news documents of text based on the refute claims as prompts.

\begin{figure}[H]
\centering       \includegraphics[width=\columnwidth]{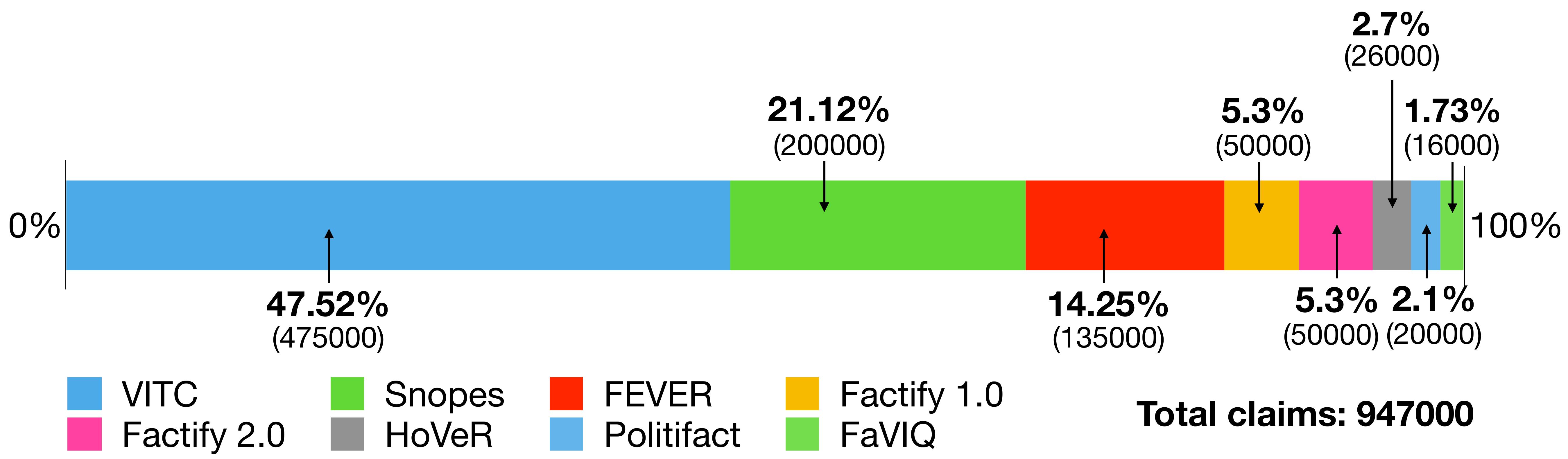}
\caption{Distribution of our dataset delineating its constituent components. }
\label{fig:datasets}
\end{figure}

We have adopted an automatic method to compile a list of claims for \textit{support} and \textit{neutral} categories. It is often seen that the same event gets reported by two different media houses separately on the same day - therefore, one can be treated as support for the other. With this strategy in mind, we have collected and annotated large-scale data automatically (c.f. appendix \ref{sec:app-A} for details). Fig. ~\ref{fig:datasets} visualizes how much data from each dataset are compiled.

Table \ref{tab:glance} offers a statistical description of the five entailment classes, their definitions in terms of textual/visual support, an approximate count of the claims, paraphrased claims, images, 5W QA pairs, evidence documents, and adversarial stories for the \texttt{Refute} class.  Furthermore, to solidify the idea behind the above categories, fig. \ref{fig:examples} offers a walk-through of an example from the FACTIFY 3M dataset.

There are only a handful of other previous works, \cite{yao2022end}, \cite{abdelnabi2022open}, \cite{9533916}, \cite{nielsen2022mumin}, \cite{jin2017multimodal}, \cite{luo2021newsclippings}, that have discussed multimodal fact verification. None of them generated large resources like ours and did not discuss QA-based explanation, heatmap-based image explainability, and adversarial assertion.

\section{Paraphrasing textual claims}\label{sec:par-text-claims}

A claim may have multiple diverse manifestations depending on the style and manner in which it was reported. Specifically, the textual component (i.e., prose) may have variations as highlighted in fig. \ref{fig:claim_venice}. We seek to echo such heterogeneity to ensure the real-world applicability of our benchmark (c.f. examples in fig. \ref{fig:chatgpt-image} ).
Manual generation of possible paraphrases is undoubtedly ideal but is time-consuming and labor-intensive. On the other hand, automatic paraphrasing has received significant attention in recent times \cite{paraphrase_1, paraphrase_2, paraphrase_3, paraphrase_4, paraphrase_5}. Our criteria for selecting the most appropriate paraphrasing model was the linguistic correctness of the paraphrased output and the number of paraphrase variations. To achieve this, we propose the following process - let's say we have a claim $c$, we generate a set of paraphrases of $c$.
Textual paraphrase detection is a well-studied paradigm, there are much state-of-the-art (SoTA) systems \cite{wang2021entailment, wang2019structbert, tay2021charformer}. We pick the best model available mostly trained based on the resources available from SNLI \cite{bowman2015large}. Next, we use the entailment model \cite{wang2019structbert} to choose the right paraphrase candidate from the generated set, by doing a pairwise entailment check and choosing only examples which exhibit entailment with $c$. 
We empirically validated the performance of GPT3 \cite{brown2020language}, Pegasus \cite{zhang2020pegasus}, and ChatGPT \cite{schulman2022chatgpt} models for our use-case and found that GPT3-text-davinci-003 outperformed the rest (c.f. appendix \ref{sec:app-par-text-claims} for details - we have evaluated models on three factors: (i) coverage, (ii) correctness and (iii) diversity).

\section{Visual paraphrase - Stable Diffusion based image synthesis}\label{sec:stable-diffusion}

While textual diversity in claims seen in the wild is commonplace, typically the visual components -- particularly, images -- also show diversity.
 The concept of AI-based text-to-image generators has been around for the past several years, but their outputs were rudimentary up until recently. In the past year, text prompt-based image generation has emerged in the form of DALL-E \cite{dalle}, ImageGen \cite{imagen}, and Stable Diffusion \cite{rombach2021highresolution}. While these new-age systems are significantly more powerful, they are a double-edged sword. They have shown tremendous potential in practical applications but also come with their fair share of unintended use cases. One major caution is the inadvertent misuse of such powerful systems. To further this point, we have utilized Stable Diffusion 2.0 \cite{rombach2021highresolution} to generate a large amount of fake news data.

Stable Diffusion \cite{rombach2021highresolution} is a powerful, open-source text-to-image generation model. The model is not only extremely capable of generating high-quality, accurate images to a given prompt, but this process is also far less computationally expensive than other text-conditional image synthesis approaches such as \cite{COGVIEW, GLIDE, lafite, make-a-scene}. Stable diffusion works on stabilizing the latent diffusion process \act{which has an aspect of randomization}, as a result, it generates a different result each time. Moreover, quality control is a challenge. We have generated 5 images for a given claim and then further ranked them, discussed in the next section.

\subsection{Re-ranking of generated images}
In order to quantitatively assess and rank the images generated by the stable diffusion model, we leverage the CLIP model \cite{radford2021learning} to obtain the best image conditioned on the prompt. We use CLIP-Score based re-ranking to select the best image corresponding to the prompt. The CLIP-Score denotes the proximity between the final image encodings and the input prompt encoding.

\subsection{Pixel-level image heatmap}\label{subsec:DAAM}
\cite{tang2022daam} perform a
text\textendash image attribution analysis on Stable Diffusion. To produce pixel-level
attribution maps, authors propose Diffusion Attentive Attribution Maps (DAAM), a novel interpretability
method based on upscaling and aggregating cross-attention
activations in the latent denoising subnetwork. We adapt the official code available on Github \cite{castorini_2022} to obtain the attribution maps in the generated images for each word in the cleaned prompt (pre-processed prompt after removal of stop-words, links, etc.). See fig. \ref{fig:daam_venice} for examples.

\subsection{Assessing quality of synthetically generated images}\label{sd-assess}

While SD has received great acclaim owing to its stellar performance for a variety of use cases, to our knowledge, to our knowledge, we are the first to adopt it for fake news generation. As such, to assess the quality of generated images in the context of the fake news generation task, we utilize two evaluation metrics. 

We use Fréchet inception distance (FID) \cite{fid} which captures both fidelity and diversity and has been the de-facto standard metric for SoTA generative models \cite{fid_paper_1, fid_paper_2, fid_paper_3, brock2018large}. The process we adopted to compute the FID score to quantitatively assess the quality of SD-generated images is detailed in \ref{quant_measures}. For a chosen set of $500$ claims (100 per category), we obtained an FID score. We obtained an FID score of $8.67$ (lower is better) between the set of real images and SD-generated images for the same textual claims.

\begin{figure}[H]
\center
\includegraphics[width=0.70\columnwidth]{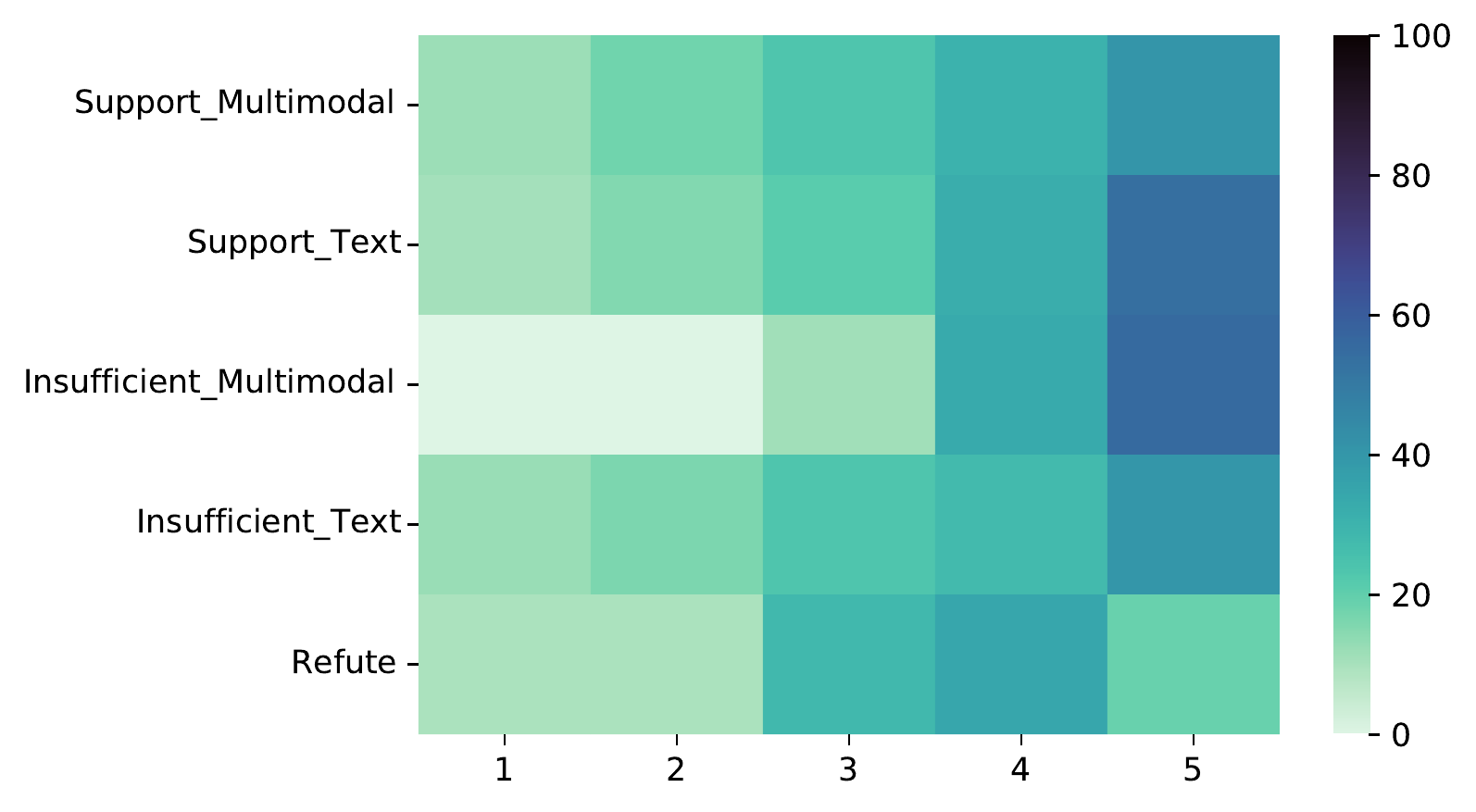}
\caption{Heatmap of MOS scores with $500$ assessed samples for each category.}
\label{fig:mos}
\end{figure}

As our secondary metric, we utilized Mean Opinion Score (MOS) at the claim category level which is a numerical measure of the human-judged perceived quality of artificially generated media (c.f. appendix \ref{apx-mos} for process details). Results of the conducted MOS tests are summarized in fig. \ref{fig:mos}.

\section{Automatic 5W QA pair generation}\label{sec:5WQA}
A false claim is very likely to have some truth in it, some correct information. In fact, most fake news articles are challenging to detect precisely because they are mostly based on correct information, deviating from the facts only in a few aspects. That is, the misinformation in the claim comes from a very specific, inaccurate statement. So, given our textual claim and image claim, we generate 5W question-answer pairs by doing semantic role labeling on the given claim. The task is now based on the generated QA pairs; a fact-checking system can extract evidence sentences from existing authentic resources to verify or refute the claim based on each question- \textit{Who, What, When, Where, and Why}.

\textls[-5]{We leverage the method of using 5W SRL to generate QA pairs \cite{rani2023factify} and verify each aspect separately to detect `\textit{exactly where the lie lies}'. This, in turn, provides an explanation of why a particular claim is refutable since we can identify exactly which part of the claim is false.}

\subsection{5W semantic role labelling}
Identification of the functional semantic roles played by various words or phrases in a given sentence is known as semantic role labeling (SRL). SRL is a well-explored area within the NLP community. There are quite a few off-the-shelf tools available: (i) Stanford SRL \cite{conf/acl/ManningSBFBM14}, (ii) AllenNLP \cite{allennlpsrl}, etc. A typical SRL system first identifies verbs in a given sentence  and then marks all the related words/phrases haven relational projection with the verb and assigns appropriate roles. Thematic roles are generally marked by standard roles defined by the Proposition Bank (generally referred to as PropBank) \cite{palmer2005proposition}, such as: \textit{Arg0, Arg1, Arg2}, and so on. We propose a mapping mechanism to map these PropBank arguments to 5W semantic roles. The conversion table \ref{tab:deppar-SRL} and necessary discussion can be found in appendix \ref{sec:app-5W}.

\subsection{Automatic 5W QA pair generation}
\begin{figure}[!ht]
    \centering
    \includegraphics[width=\columnwidth]{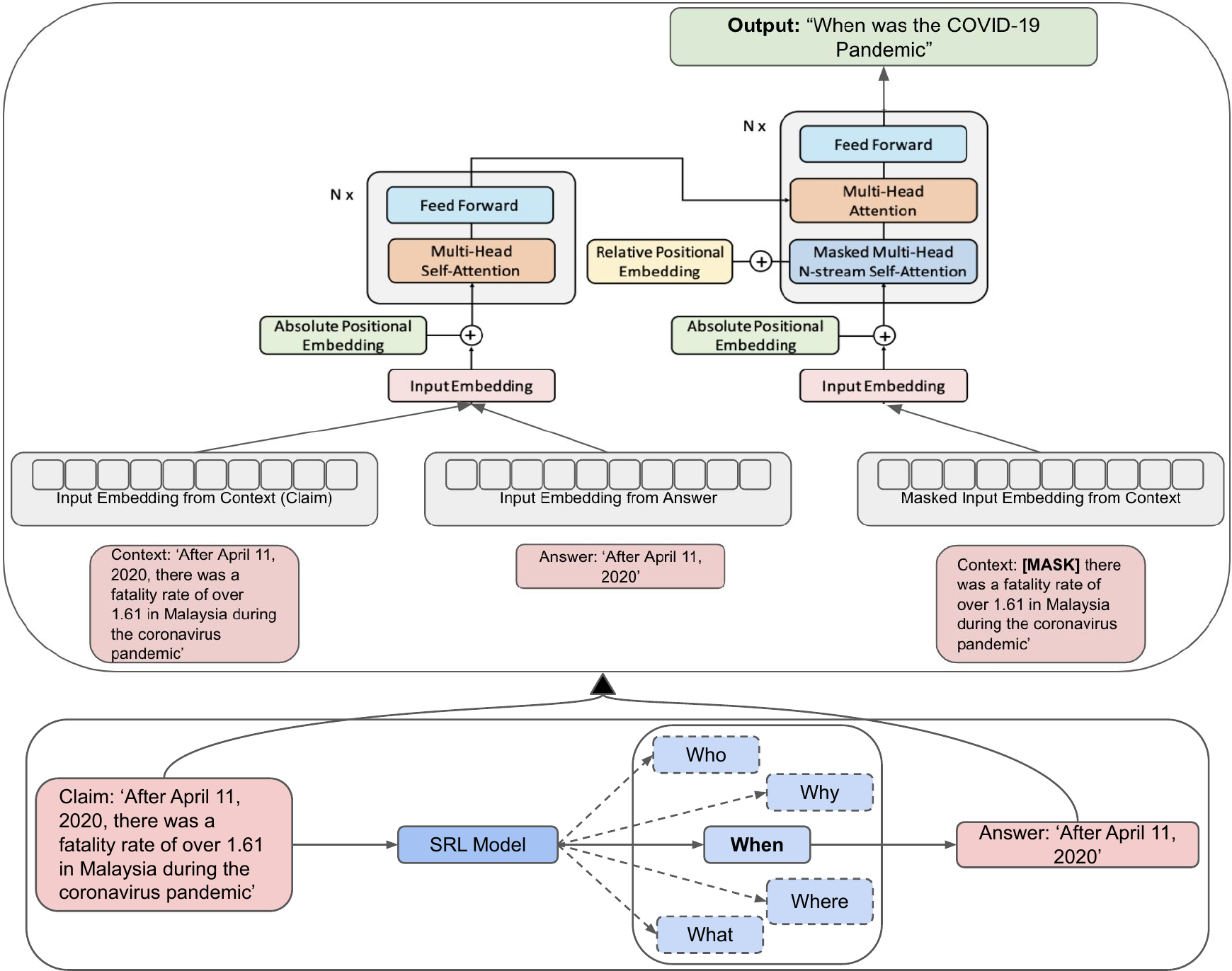}
    \caption{Illustration of 5W QA Generation Pipeline using ProphetNet.}
    \label{fig:5w_qa_gen}
\end{figure}

We present a system for generating 5W aspect-based questions generation using a language model (LM) 
that is fed claims as input and uses the SRL outputs as replies to produce 5W questions with respect to the 5W outputs. We experimented with a variety of LMs: BART \cite{lewis2019bart} and ProphetNet \cite{qi2020prophetnet}, eventually settling on ProphetNet (see fig. ~\ref{fig:5w_qa_gen}) based on empirically figuring out the best fit for our use-case (c.f. appendix \ref{subsec:app-CA} for details).

We then create answers using the evidence from the questions generated using ProphetNet by running them through T5 \cite{yamada2020luke} -- a SoTA QA model, using the 5W-based generated questions. See section ~\ref{sec:5w_qa_val} for details.

\section{Injecting adversarial assertion}\label{sec:adv}
The rise of generative AI techniques with capabilities that mimic a human's creative thought process has led to the availability of extraordinary skills at the masses' fingertips. 
This has led to the proliferation of generated content, which is virtually indistinguishable as real or fake to the human eye, even for experts in some cases. This poses unparalleled challenges to machines in assessing the veracity of such content. 

As one of the novelties of our work, we address this by introducing synthetically generated adversarial fake news documents for all the \textit{refute} claims using OPT \cite{zhang2022opt}, a large language model. In doing so, we attempt to confuse the fact verification system by injecting fake examples acting as an adversarial attack. To draw a parallel in a real-world scenario, this could mean the proliferation of such fake news articles online via social media, blog posts, etc. which would eventually lead to a fact-verification system being unable to make a concrete decision on the trustworthiness of the news. Such a scenario would lend itself as a natural manifestation of an adversarial attack by virtue (rather, the "vice") of the fake news articles confusing the fact verification system. We analyze the impact on the performance of our fact verification system in table \ref{table:baseline-results}. Our goal in offering these adversarial articles as one of our contributions is to provide future researchers a benchmark using which they can measure (and hence, improve) the performance of their fact verification system.

\subsection{Accuracy of text generation}
We assess the quality of text generation using the following evaluation metrics:

\textbf{Fluency:} We measure the text fluency using perplexity scores. The perplexity is a measure of the likelihood of the generated sentence on a language model. We use a pre-trained GPT-2 model to evaluate text perplexity. A lower value is preferred. We have used the GPTZero detector to evaluate our perplexity score \cite{tian_2023}. Checking for paraphrased text generated over a 50 claims (25 original and 25 adversarial), we report an average perplexity score of 129.06 for original claims and 175.80 for adversarial claims (more details in appendix ~\ref{subsec:app-adv-assertion}). We intend to perform more evaluations on standard fluency measures such as ROGUE \cite{rouge} and BLEURT \cite{bleurt}.

\section{Experiments - baselines \& performance} \label{sec:experiments}

In this section, we present baselines for: \textit{(i) multimodal entailment, (ii) 5W QA-based validation, and (iii) results of our models after adversarial injections of generated fake news stories}. 

\begin{table*}[t!]
    \centering
\resizebox{\textwidth}{!}{%
    \begin{tabular}{ccccccccccccc}
     \toprule
    & \multicolumn{2}{c}{\texttt{Support\_Text}} & 
\multicolumn{2}{c}{\texttt{Support\_Multimodal}} & \multicolumn{2}{c}{\texttt{Insufficient\_Text}} & \multicolumn{2}{c}{\texttt{Insufficient\_Multimodal}} & \multicolumn{2}{c}{\texttt{Refute}} & 
\multicolumn{2}{c}{\texttt{\textbf{Average}}}\\
      & Text-only & Multimodal & Text-only & Multimodal & Text-only & Multimodal & Text-only & Multimodal & Text-only & Multimodal & Text-only & Multimodal \\
     \midrule
Pre-adversarial attack (F1) & 
\multicolumn{2}{c}{\resizebox{160pt}{!}{\includegraphics[trim = {0.5cm 0.25cm 0cm 0cm}]{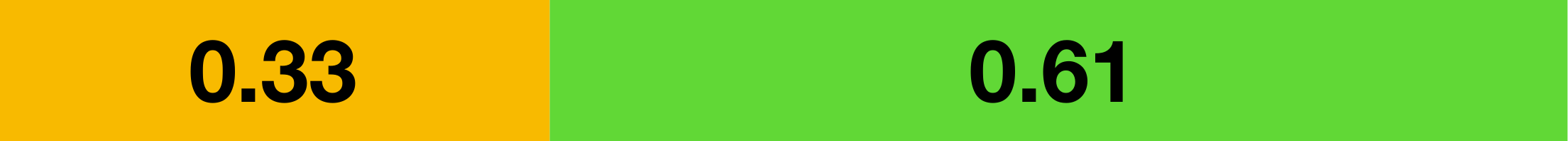}}} & 
\multicolumn{2}{c}{\resizebox{160pt}{!}{\includegraphics[trim = {0.5cm 0.25cm 0cm 0cm}]{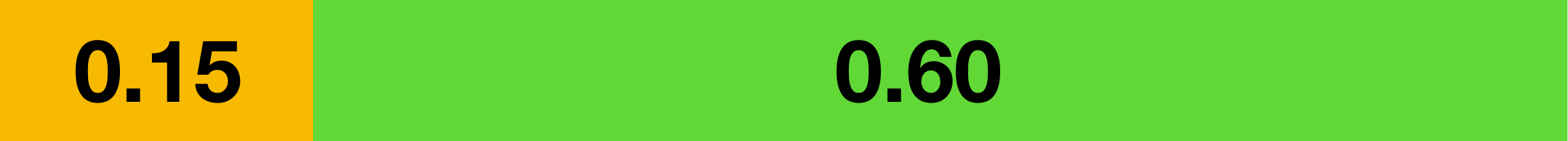}}} & 
\multicolumn{2}{c}{\resizebox{160pt}{!}{\includegraphics[trim = {0.5cm 0.25cm 0cm 0cm}]{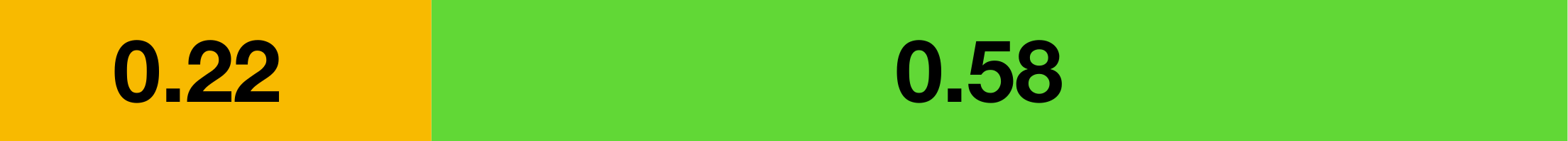}}} & 
\multicolumn{2}{c}{\resizebox{160pt}{!}{\includegraphics[trim = {0.5cm 0.25cm 0cm 0cm}]{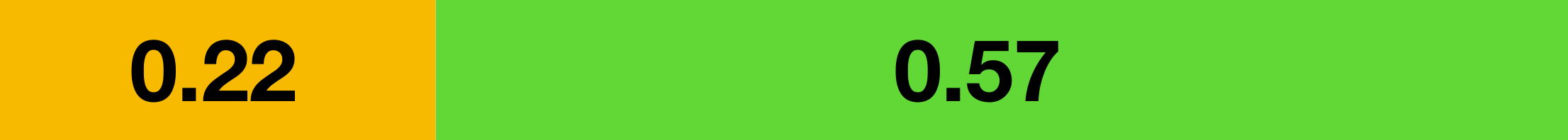}}} & 
\multicolumn{2}{c}{\resizebox{160pt}{!}{\includegraphics[trim = {0.5cm 0.25cm 0cm 0cm}]{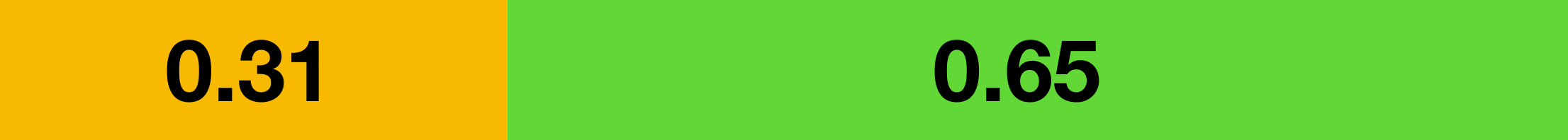}}} & 
\multicolumn{2}{c}{\resizebox{160pt}{!}{\includegraphics[trim = {0.5cm 0.25cm 0cm 0cm}]{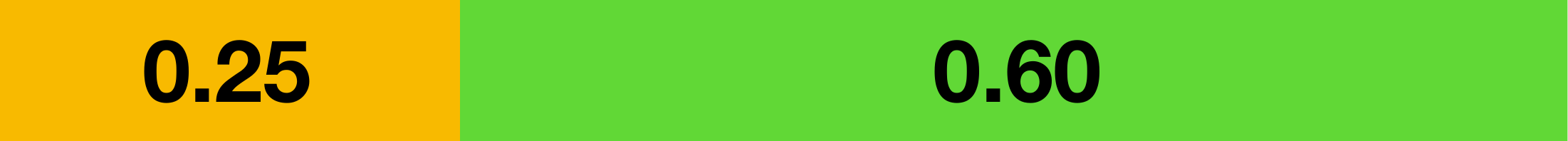}}} \\
\hline
Post-adversarial attack (F1) & 
\multicolumn{2}{c}{\resizebox{160pt}{!}{\includegraphics[trim = {0.5cm 0.5cm 0cm 0cm}]{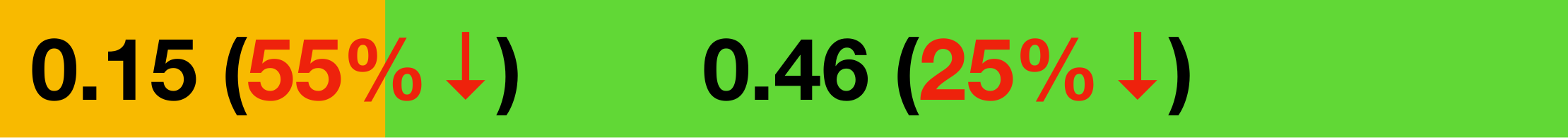}}} & 
\multicolumn{2}{c}{\resizebox{160pt}{!}{\includegraphics[trim = {0.5cm 0.5cm 0cm 0cm}]{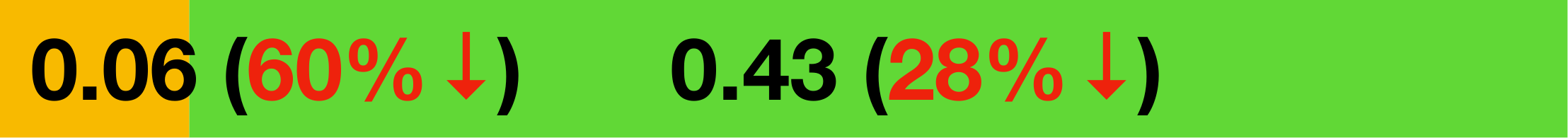}}} & 
\multicolumn{2}{c}{\resizebox{160pt}{!}{\includegraphics[trim = {0.5cm 0.5cm 0cm 0cm}]{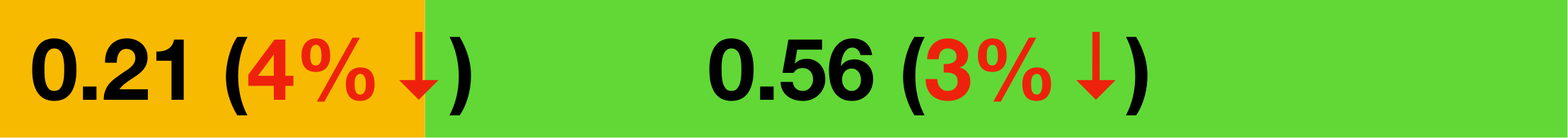}}} & 
\multicolumn{2}{c}{\resizebox{160pt}{!}{\includegraphics[trim = {0.5cm 0.5cm 0cm 0cm}]{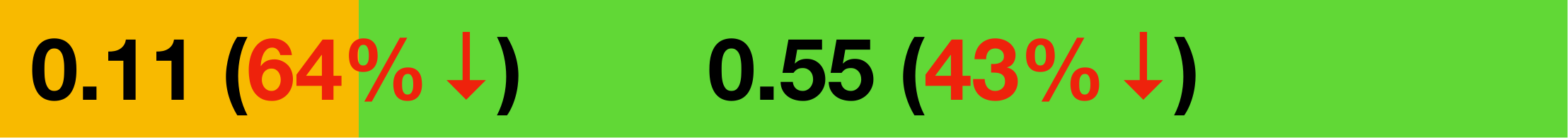}}} & 
\multicolumn{2}{c}{\resizebox{160pt}{!}{\includegraphics[trim = {0.5cm 0.5cm 0cm 0cm}]{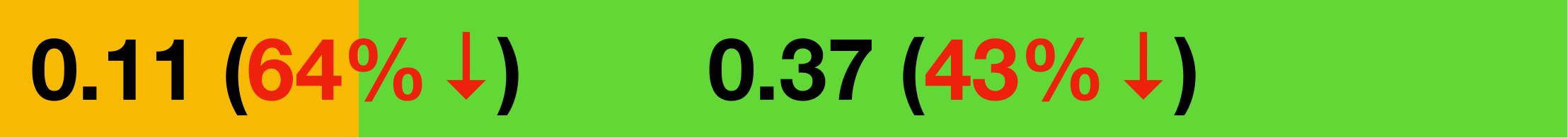}}} & 
\multicolumn{2}{c}{\resizebox{160pt}{!}{\includegraphics[trim = {0.5cm 0.5cm 0cm 0cm}]{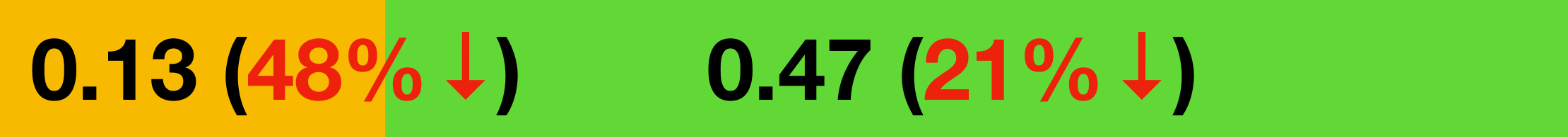}}} \\      
\bottomrule
    \end{tabular}%
    }
    \caption{Results of the text-only and multimodal baselines pre- and post-adversarial attack.}
    \label{table:baseline-results}
\end{table*}

\subsection{Multimodal entailment: \textit{support or refute?}}
In this paper, we model the task of detecting multimodal fake news as multimodal entailment. We assume
that each data point contains a reliable source of information, called \textit{document}, and its
associated image and another source whose validity must be assessed, called the \textit{claim} which
also contains a respective image. The goal is to identify if the claim entails the document.
Since we are interested in a multimodal scenario with both image and text, entailment has two
verticals, namely textual entailment, and visual entailment, and their respective combinations.
This data format is a stepping stone for the fact-checking problem where we have one reliable
source of news and want to identify the fake/real claims given a large set of multimodal claims.
Therefore the task essentially is: given a textual claim, claim image, text document, and document
image, the system has to classify the data sample into one of the five categories: \texttt{Support\_Text},
\texttt{Support\_Multimodal}, \texttt{Insufficient\_Text}, \texttt{Insufficient\_Multimodal}, and \texttt{Refute}. Using the Google Cloud Vision API \cite{visionapi}, we also perform OCR to obtain the text embedded in images and utilize that as an additional input.

\begin{figure}[H]
\centering
\includegraphics[width=0.85\columnwidth, trim = {0.3cm 0.65cm 0 0}]{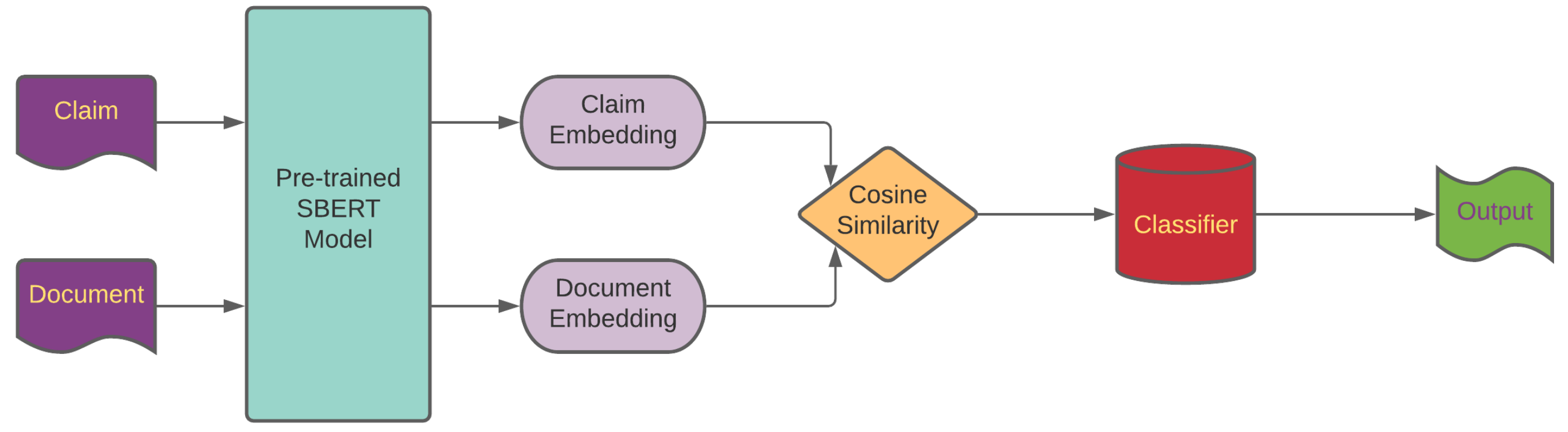}

\caption{Text-only baseline model which takes only claim text and document text as input.}
\label{fig:Text-only-baseline-factify}
\end{figure}

\textbf{Text-only model:} Fig. \ref{fig:Text-only-baseline-factify} shows our text-only model, which adopts a siamese architecture focussing only on the textual aspect of the data and ignores the visual information. To this end, we generate sentence embeddings of the claim and document attributes using a pretrained MPNet Sentence BERT model \cite{sentence-bert} (specifically the \texttt{all-mpnet-base-v2} variant). Next, we measure the cosine similarity using the generated embeddings. The score, thus generated, is used as the only feature for the dataset, and classification is evaluated based on their F1 scores. 

Table \ref{table:baseline-results} shows the F1 score for the unimodal (i.e., text-only) and multimodal approaches (c.f. pre-adversarial attack row in table \ref{table:baseline-results}) trained using their respective feature sets. The multimodal model shows a distinct improvement in performance compared to the text-only model, indicating the value-addition of the visual modality. 

\begin{figure}[h]
\centering
\includegraphics[width=0.85\columnwidth, trim = {0.15cm 0.8cm 0 0}]{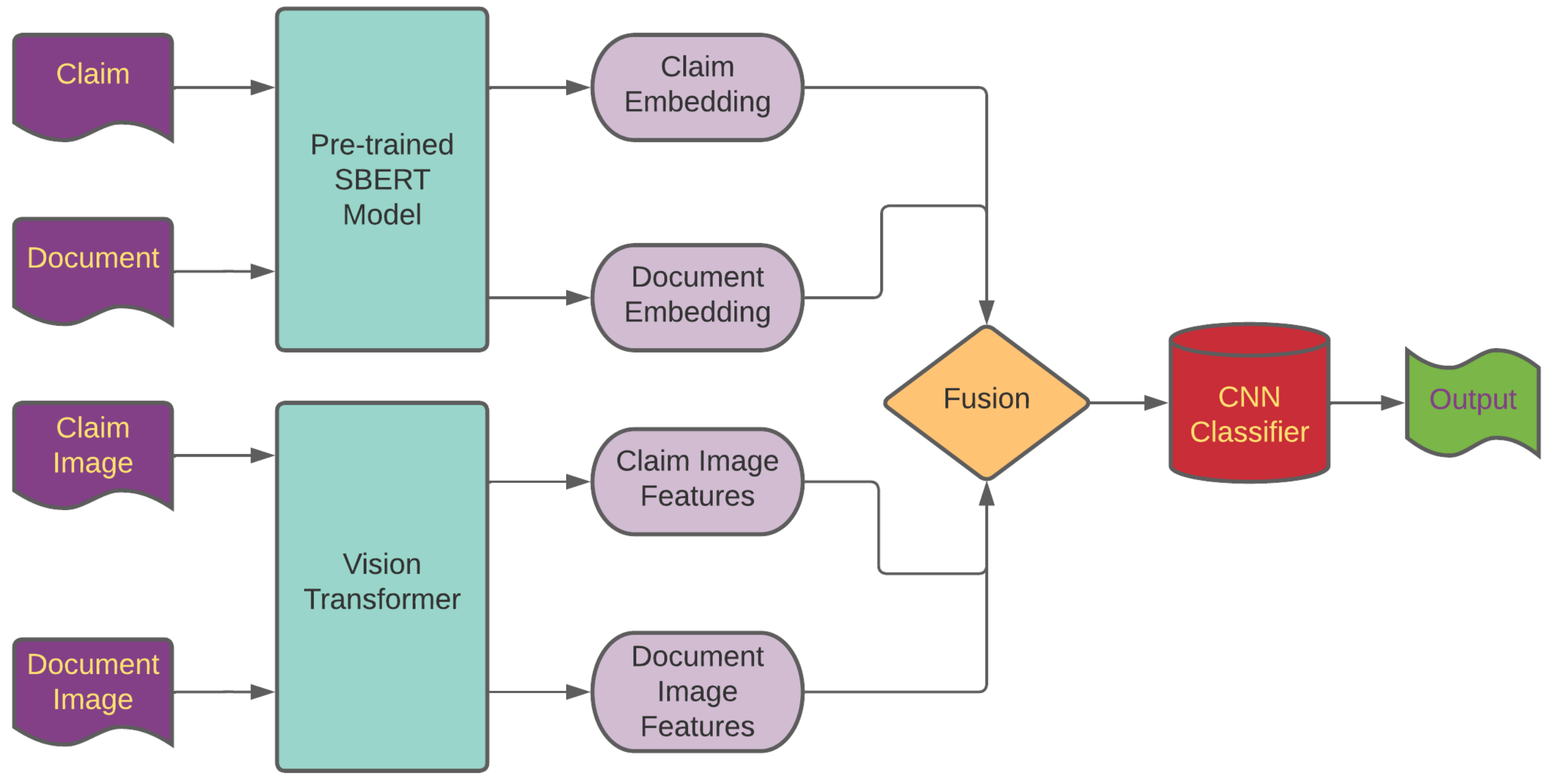}
\caption{Multimodal baseline model which takes as input: (i) claim text, (ii) claim image, (iii) document text, and (iv) document image.}
\label{fig:Factify2_Baseline}
\end{figure}

\subsection{ 5W QA-based validation}
\label{sec:5w_qa_val}
We generate 5W Question-Answer pairs for the claims, thus providing explainability along with evidence in the form of answers generated. To this end, we use the SoTA T5 \cite{raffel2020exploring} model for question answering (QA) in this work. It is trained using a modified version of BERT's masked language model, which involves predicting masked words and entities in an entity-annotated corpus from Wikipedia.

\begin{figure}[h!]
\center
\includegraphics[width=0.8\columnwidth, trim = {0.15cm 0.6cm 0 0}]{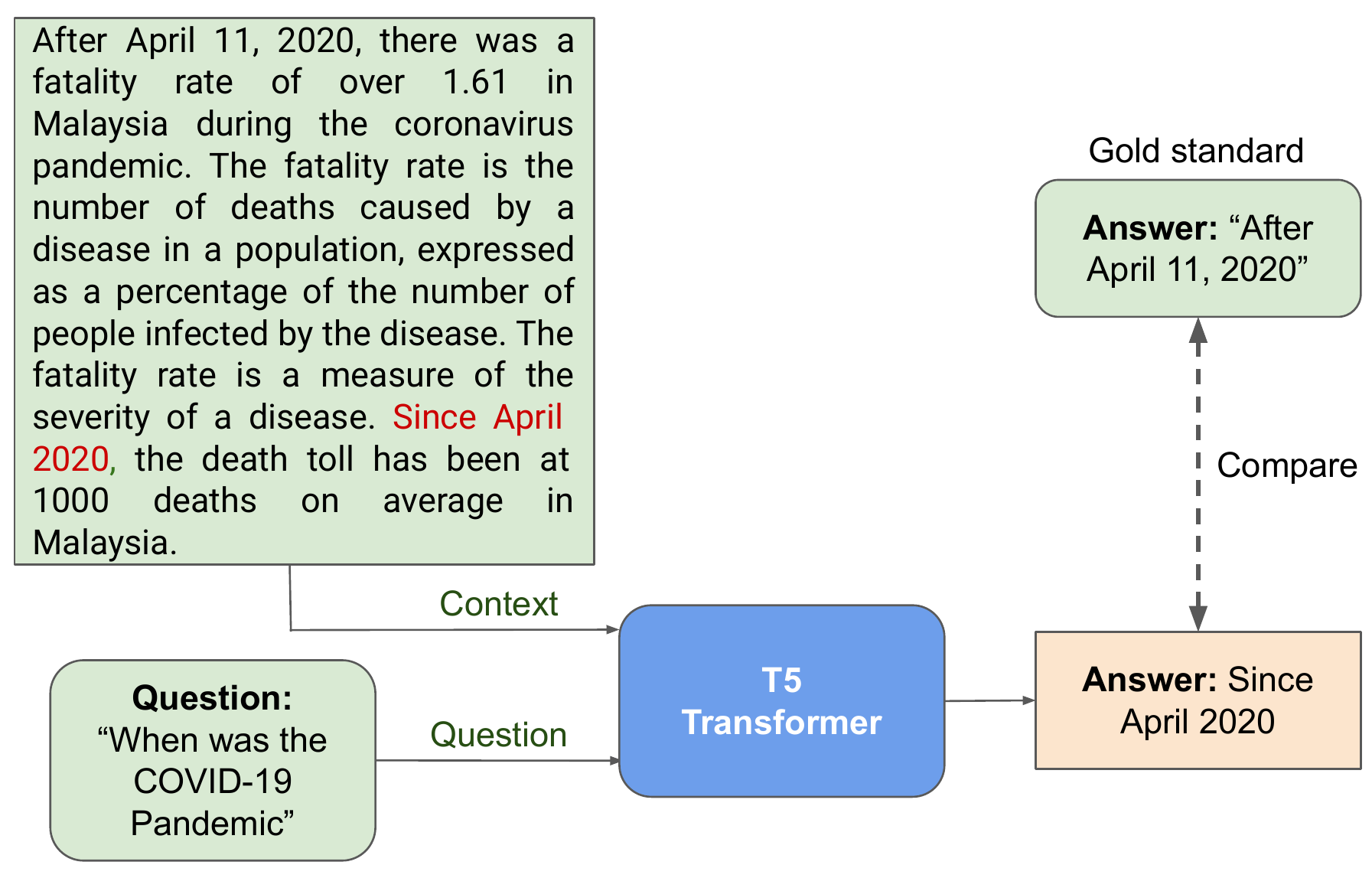}
\caption{T5-based question answering framework.}
\label{fig:Question answering Framework}
\end{figure}

\subsection{Adversarial attack}
With the emergence of prolific ChatGPT, the risk of AI-generated content has reached an alarming apocalypse. ChatGPT has been declared banned by the school system in NYC \cite{rosenblatt_2023}, Google ads \cite{grant_metz_2022}, and Stack Overflow \cite{makyen_olson_1969}, while scientific conferences like ACL \cite{chairs_2023} and ICML \cite{foundation_2023} have released new policies deterring the usage of ChatGPT for scientific writing. Indeed, the detection of AI-generated text has suddenly emerged as a concern that needs imminent attention. While watermarking as a potential solution to the problem is being studied by OpenAI \cite{tc-gpt}, a handful of systems that detect AI-generated text such as GPT-2 output detector \cite{gpt-od}, GLTR \cite{gltr}, GPTZero \cite{GPTZero}, has recently been seen in the wild. Furthermore, these tools typically only produce meaningful output after a minimum (usually, 50+) number of tokens. We tested GPTZero on a randomly selected set of $100$ adversarial samples, equally divided into human-generated text and AI-generated text. Our results indicate that these systems are still in their infancy (with a meager 22\% accuracy). It is inevitable that AI-generated text detection techniques such as watermarking, perplexity, etc. will emerge as important paradigms in generative AI in the near future, and FACTIFY 3M will serve the community as a benchmark in order to test such techniques for fact verification.  

Table \ref{table:baseline-results} shows the F1 score post adversarial attack for the unimodal (i.e., text-only) and multimodal approaches - proving that injecting adversarial news can confuse fact-checking very easily.

\section{Conclusion and future avenues}\label{sec:conclusion}
We are introducing FACTIFY 3M, the largest dataset and benchmark for multimodal fact verification. We hope that our dataset facilitates research on multimodal fact verification on several aspects - \textit{(i) visual QA-based explanation of facts, (ii) how to handle adversarial attacks for fact verifications, (iii) whether generated images can be detected, and (iv) 5W QA-based  help journalists to fact verify easily for complex facts}. FACTIFY 3M will be made public and open for research purposes.

\newpage
\section{Discussion and limitations}
\label{sec:limitations}
In this section, we mention a few aspects that could be improved and also detail how we plan to improve upon those specific aspects.

\subsection{Paraphrasing claims}
\label{sec:limitations_paraphrase}
Manual generation of possible paraphrases is undoubtedly ideal but is time-consuming and labor-intensive. Automatic paraphrasing is a good way to scale quickly, but there could be more complex variations of meaning paraphrases hard to generate automatically. For example - "\textit{It's all about business - a patent infringement case against Pfizer by a rival corporate reveals they knew about COVID in one way!}" and "\textit{Oh my god COVID is not enough now we have to deal with HIV blood in the name of charity!}". 

An ideal for this shortcoming would be to manually generate a few thousand paraphrase samples and then fine-tune language models. On the other hand, a new paradigm in-context Learning is gaining momentum \cite{7580601}. In-context learning has been magical in adapting a language model to new tasks through just a few demonstration examples without doing gradient descent. There are quite a few recent studies that demonstrate new abilities of language models that learn from a handful of examples in the context (in-context learning - ICL for short). Many studies have shown that LLMs can perform a series of complex tasks with ICL, such as solving mathematical reasoning problems \cite{https://doi.org/10.48550/arxiv.2201.11903}. These strong abilities have been widely verified as emerging abilities for large language models \cite{https://doi.org/10.48550/arxiv.2201.11903}. From prompt engineering to chain of thoughts, we are excited to do more experiments with the new paradigm of in-context learning for automatically paraphrasing claims.

\subsection{Image synthesis using Stable Diffusion}
\label{sec:limitations_sd}
Although, in general, the quality of the image synthesized by Stable Diffusion is great, it does not perform well in two cases - i) \textit{very long text} (more than 30 words or so, multiple sentence claim, etc.), ii) \textit{text with metaphoric twists} - for example, "\textit{It's all about business - a patent infringement case against Pfizer by a rival corporate reveals they knew about COVID in one way!}" and "\textit{Oh my god COVID is not enough now we have to deal with HIV blood in the name of charity!}". It is worthy seeing how in-domain adaptation could be made for SD image synthesis, inspired from \cite{ruiz2022dreambooth}.

\subsection{5W SRL}
\label{sec:limitations_5wsrl}
Semantic role labeling is a well-studied  sub-discipline, and the mapping mechanism we proposed works well in most cases except in elliptic situations like anaphora and cataphora. In the future, we would like to explore how an anaphora and coreference resolution \cite{joshi2019coref} can aid an improvement.

\subsection{5W QA pair generation}
\label{sec:limitations_qg}
5W semantic role-based question generation is one of the major contributions of this paper. While automatic generation aided in scaling up the QA pair generation, it also comes with limitations of generating more complex questions covering multiple Ws and \textit{how} kinds of questions. For example - "\textit{How Moderna is going to get benefited if this Pfizer COVID news turns out to be a rumor?}". For the betterment of FACTIFY benchmark, we would like to generate few thousand manually generated abstract QA pairs. Then will proceed 
towards in-context Learning \cite{7580601}.

Abstractive question-answering has received momentum \cite{zhao-etal-2022-compositional}, \cite{pal-etal-2022-parameter} recently. We want to explore how we can generate more abstract QA pairs for the multimodal fact-verification task. 

\subsection{QA system for the 5W question}
\label{sec:limitations_qa}
Generated performance measures attest the proposed QA model needs a lot more improvement. This is due to the complexity of the problem and we believe that will attract future researchers to try this benchmark and conduct research on multimodal fact verification. 

It has been realized by the community that relevant document retrieval is the major bottleneck for fact verification. Recent work introduced a fresh perspective to the problem - named Hypothetical Document Embeddings (HyDE) \cite{hyde} and applied a clever trick even if the wrong answer is more semantically similar to the right answer than the question. This could be an interesting direction to explore and examine how that could aid in retrieving relevant documents and answers. 

\subsection{Adversarial attack}
\label{sec:limitations_adverserial}
Precisely, we are the first to formally introduce an adversarial attack for fact verification and introducing large-scale data. While it is a hot topic of discussion how systems can identify AI-generated text, there is no breakthrough so far. We would like to explore more in this direction more, specifically for multimodal fact verification.

\bibliography{ref}
\bibliographystyle{acl_natbib}

\newpage
\onecolumn
\section*{Frequently Asked Questions - FAQs}\label{sec:FAQs}

\begin{enumerate}
    \item Does Stable Diffusion offer the adeptness to generalize and scale to different real-world scenarios? In other words, Stable Diffusion is great at generating one-off plausible examples but is generalizability to life's combinatorial scenarios a concern?
    \begin{description}
    \item \textbf{Ans.} - Fake news is generally written connecting popular topics and personalities, therefore stable diffusion does a decent job. However, there are some limitations, which we discussed in detail in the limitation section ~\ref{sec:limitations_sd}. 
    Moreover, we have generated stable diffusion images for all claims as a visual paraphraser as mentioned in section \ref{sec:stable-diffusion}.
    Furthermore, we have presented a holistic evaluation through objective (FID) and subjective (MOS) metrics. Please refer to the section \ref{sd-assess}, table ~\ref{fig:mos}.
    \end{description}
    \item What are the novel assertions in this paper if the multimodal data is generated automatically using SoTA generative models?
    \begin{description}
    \item \textbf{Ans.} - The novelty of this work is three-fold: 
    \begin{enumerate}
        \item Justification of the classification using 5WQA verification.
        \item Injecting an adversarial attack in the form of fake news to make the dataset more robust.
        \item Adding synthetically generated images using Stable Diffusion to enhance multimodal data.
    \end{enumerate}
    \end{description}
    \item 5W SRL is understandable, but how is the quality of the 5W QA pair generation using a language model?
    \begin{description}
    \item \textbf{Ans.} - We have evaluated our QA generation against the SoTA model for QA Tasks - T5. Please refer to the appendix section \ref{sec:app-5W valn}, table \ref{tab:QAG-QA} for a detailed description of the process and evaluation. Moreover, please see the discussion in the limitation section ~\ref{sec:limitations_qg}.
    \end{description}
    \item What is the overarching idea we're trying to highlight by introducing an adversarial attack? 
    \begin{description}
    \item \textbf{Ans.} - The broader point that the introduction of an adversarial attack indicates is that a fact verification model needs to be more robust in combating synthetically generated fake news, which is easily publishable by wrongdoers on the internet. This is of extreme relevance today as AI-assisted writing has become very popular and miscreants spread fake news taking advantage of LLMs. 
    \end{description}
    \item How does adversarial attack impact the performance? 
    \begin{description}
    \item \textbf{Ans.} - As reported in table \ref{table:baseline-results}, we see that the performance of the model drops across all categories post adversarial attack using fake claims. This is seen in both instances: text-only and multimodal model. 
    \end{description}
    \item Despite the controversies surrounding AI-assisted writing, why have we still chosen to use LLMs as our paraphrasers?
    \begin{description}
    \item \textbf{Ans.} - The controversy lies mostly in a conversational setting or creative writing. When it comes to paraphrasing news claims, we have empirically found that GPT-3 (specifically the \texttt{text-davinci-003} variant) \cite{brown2020language} performs better in comparison to other models such as Pegasus \cite{zhang2020pegasus}, and ChatGPT \cite{schulman2022chatgpt}.
    \end{description}
    \item What was the chosen metric of evaluation for text generation using LLMs?
    \begin{description}
    \item \textbf{Ans.} - For now, we have evaluated adversarial claims using GPTZero text detector. Evaluation on standard metrics such as control and fluency will be made public along with our dataset.
    \end{description}
\end{enumerate}

\newpage
\onecolumn

\appendix
\renewcommand{\thesubsection}{\Alph{section}.\arabic{subsection}}
\renewcommand{\thesection}{\Alph{section}}
\setcounter{section}{0}

\section*{Appendix}\label{sec:appendix}

This section provides supplementary material in the form of additional examples, implementation details, etc. \textbf{To bolster the reader's understanding of the concepts presented in this work.}

\section{Data sources and compilation}\label{sec:app-A}
 
In this section, we provide additional details on data collection and compilation. As mentioned in section ~\ref{sec:related} we are only interested in refute category from the available datasets, for support and neutral categories we have collected a significant amount of data from the web. This data collection process is semi-automatic.

For FEVER and VITC, only the claims belonging in the train split were used for making the dataset. FaVIQ \cite{park2021faviq} has two sets: Set A and Set R. Set A consists of ambiguous questions and their disambiguations. Set R is made of unambiguous question-answer pairs. We have used claims from set A in our dataset to make the entailment task more challenging. In the case of HoVer \cite{jiang2020hover}, we have used all 26171 claims for our dataset.

In the Factify dataset \cite{mishra2022factify}, the authors have collected date-wise tweets from Twitter handles of Indian and US news sources: (i) Hindustan Times \cite{times}, ANI \cite{international} for India, and (ii) ABC \cite{news}, CNN \cite{network} for the US, based on accessibility, popularity and posts per day. We drew our motivation from \cite{mishra2022factify}. Moreover, these Twitter handles are eminent for their objective and disinterested approach. From each tweet, the tweet text and the tweet image(s) have been extracted. Listing \ref{list-1} delineates each attribute in the dataset and its respective description while listing B elaborates on the process we followed for collecting data for \texttt{Support} and \texttt{Neutral} categories.

\section{Text and image similarity measures}

Table \ref{tab:glance} explains the five classes in the dataset. For the appropriate classification of the dataset, two similarity measures were computed.

\subsection{Sentence comparison} 

We adopt two methods to check similarity given a set of two sentences: 

\begin{itemize}
\item \textbf{Sentence BERT:} Sentence BERT \cite{reimers2019sentence} is a modification of the BERT model that uses a contrastive loss with a siamese network architecture to derive sentence embeddings. These sentence embeddings can be compared with each other to get their corresponding similarity score. Authors use cosine similarity as the textual similarity metric. 
\item{We utilize Sentence BERT (SBERT) \cite{reimers2019sentence} instead of alternatives such as BERT or RoBERTa, owing to its rich sentence embeddings yielding superior performance while being much more time-efficient (in terms of sentences/sec) \cite{sbert}.}
We manually decide on a threshold value \textit{T1} for cosine similarity and classify the text pair accordingly. If the cosine similarity score is greater than \textit{T1}, then it is classified into the \texttt{Support} category. On the other hand, if the cosine similarity score is lower than \textit{T1}, the news may or may not be the same (the evidence at hand is insufficient to judge whether the news is the same or not). Hence it is sent for another check before classifying it into the \texttt{Insufficient} category.
 \textbf{NLTK:} If the cosine similarity of the sentence pair is below \textit{T1}, we use the NLTK library \cite{bird2009natural} to check for common words between the two sentences.
 If the score of the common word is above a different manually decided threshold \textit{T2}, only then the news pair is classified into the \texttt{Insufficient} category. Not sure what this  sentence is trying to say - let's rephrase. Common words are being checked to ensure that the classification task is challenging. 
To check for common words, both texts in the pair are preprocessed, which included stemming and removing stopwords. The processed texts are then checked for common and similar words, and their corresponding scores are determined. If the common words score is greater than T2, the pair is classified as Insufficient else the pair is dropped. 
\end{itemize}

\begin{minipage}{0.3\textwidth}
\begin{tcolorbox}[colback=blue!5!white,colframe=blue!75!black,title={Listing A: Attributes},left=0mm,,fontupper=\linespread{0.9}\selectfont,fontlower=\linespread{0.9}\selectfont]
\begin{itemize}
\item \verb|Claim|: Tweet A text
\item \verb|Claim_image|: Tweet A image 
\item \verb|Claim_ocr|: Tweet A image OCR 
\item \verb|Document|: Tweet B article text 
\item \verb|Document_image|: Tweet B image 
\item \verb|Document_ocr|: Tweet B image OCR 
\item \verb|Category|
\end{itemize}
\label{list-1}
\end{tcolorbox}
\end{minipage}
\begin{minipage}{0.70\textwidth}
\begin{tcolorbox}[colback=blue!5!white,colframe=blue!75!black,title=Listing B: Procedure for data collection for \texttt{Support} and \texttt{Neutral} categories,halign=center,valign=center,left=0mm,fontupper=\linespread{0.9}\selectfont,fontlower=\linespread{0.95}\selectfont]
\begin{itemize}
\item For each tweet of account A, authors got similar tweets from account B. Similarity is measured on the basis of text. Text similarity is measured using Sentence BERT \cite{reimers2019sentence} first, and then the extent of common words is measured as the second metric.
\item Next, the image similarity for the corresponding images of the tweet pair was calculated. Image similarity is measured using histogram similarity and cosine similarity on a pre-trained ResNet50 model. 
\item According to the scores for each of these measures, the tweet pair is classified into 4 categories: \verb|Support_Multimodal|, \verb|Support_Text|, \verb|Insufficient_Multimodal|, and \verb|Insufficient_Text|. The various thresholds used for classification are listed in Figure \ref{fig:Classification Thresholds}.
\item From this tweet pair, authors have selected a tweet (say tweet B) and obtained the url for the corresponding article published on the source’s website from the tweet text. Then the tweet text was replaced  with article contents after scraping it (\verb|document| in dataset). This is done so as to mimic real world fact checking process, i.e., manually comparing claims with documents or articles.
\item The image OCRs were obtained using Google Cloud Vision API \cite{visionapi}. 
\vspace{-5pt}
\end{itemize}
\label{list-2}
\end{tcolorbox}
\end{minipage}

\subsection{Image comparison} 

\begin{figure}[h]
    \centering
    \includegraphics[width=13cm,  trim = {0 1.5cm 0 0}]{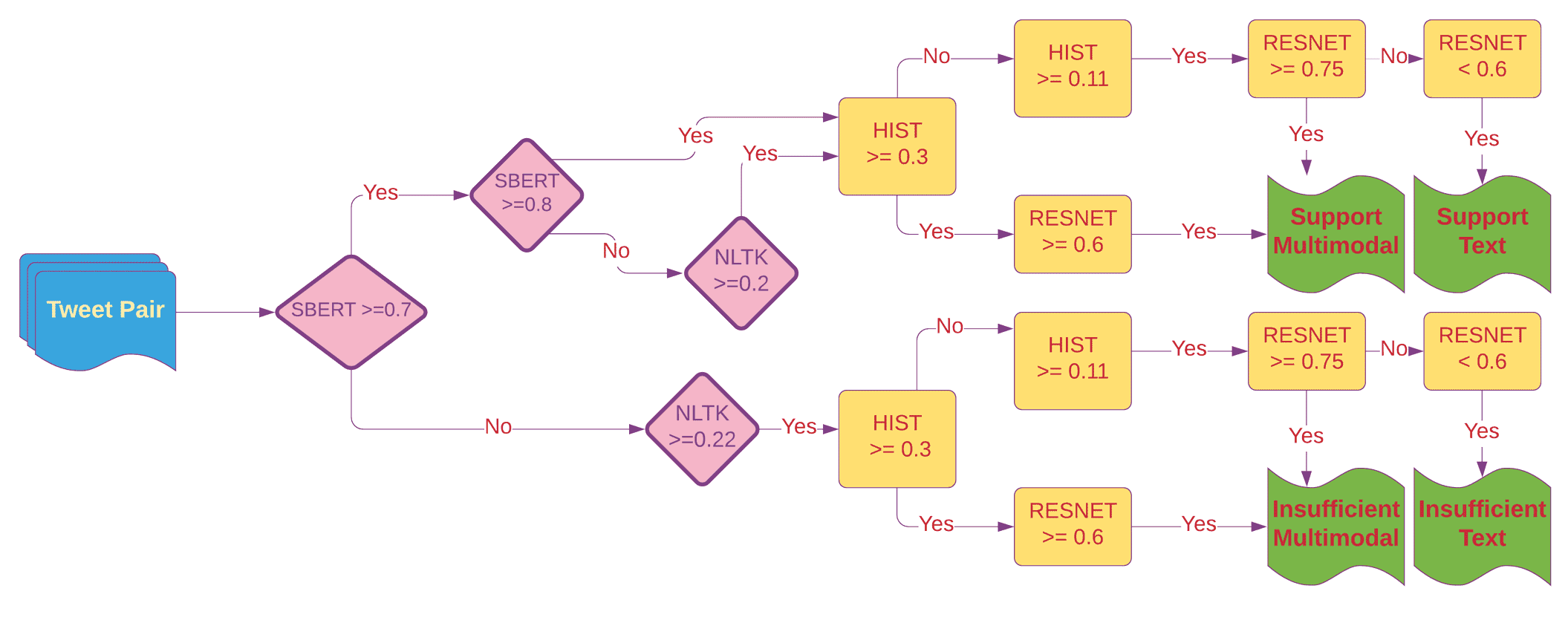}
    \caption{Text and image pair similarity based on classification thresholds on pre-trained models.}
    \label{fig:Classification Thresholds}
\end{figure}

We adopt two metrics for assessing image similarity:

\begin{itemize}
\item \textbf{Histogram Similarity:} The images are converted to normalized histogram format and similarity is measured using the correlation metric cite{}. 
\item \textbf{Cosine Similarity:} The images are converted to feature vectors using pre-trained ViT \cite{dosovitskiy2020image} model, and these feature vectors are used to calculate the cosine similarity score. 
Manually decided thresholds, as described in Figure \ref{fig:Classification Thresholds}, are used to judge whether the text and image pair is similar or not. 
\end{itemize}

The text pairs are first classified into either \texttt{Support} or \texttt{Insufficient} categories, and then  further sub-classified into \texttt{Support\_Text}/\texttt{Support\_Multimodal}, or \texttt{Insufficient\_Text}/\texttt{Insufficient\_Multimodal} categories based on the similarity of the image pairs. If the corresponding images for the texts are similar, then they could be used to judge whether news is the same or not. The category where both the images and the texts are similar is called \texttt{Support\_Multimodal}. The category where the images are similar but the texts were not is called \texttt{Insufficient\_Multimodal}. If the corresponding images for the texts were not similar, then they could not be used to judge whether news is the same or not. The category where both the images and the texts are not similar is called \texttt{Insufficient\_Text}. The category where the texts are similar but the images are not is called \texttt{Support\_Text}.

For each article published on these websites, we collect the claim (sentence that states the fake news), document (text that proves claim is false), claim images (fake news image, could be screenshot of the fake post), document image (image that is proof of the fake nature of the claim).

\section{Paraphrasing textual claims}\label{sec:app-par-text-claims}

A textual given claim may appear in various different textual forms in real life, owing to variety in the writing styles of different news publishing houses. Incorporating such variations is essential to developing a strong benchmark to ensure a holistic evaluation. This forms our motivation behind paraphrasing textual claims. Manual generation of possible paraphrases is undoubtedly ideal, but that process is time-consuming and labour-intensive. On the other hand, automatic paraphrasing has received significant attention in recent times \cite{niu2020unsupervised} \cite{zhang2020pegasus} \cite{nighojkar2021improving}. As mentioned in section \ref{sec:par-text-claims}, for a given claim, we generate multiple paraphrases using various models and perform entailment using \cite{wang2019structbert} -- a SoTA model trained on the on SNLI task \cite{bowman2015large} -- to detect how many of them are entailed in the actual claim.

In the process of choosing the appropriate model based on a list of available models, the primary question we asked is how to make sure the generated paraphrases are rich in diversity while still being linguistically correct. A top level, we delineate the process followed to achieve this as follows (more details later in this section). Let's say we have a claim $c$. We generate $n$ paraphrases using a paraphrasing model. This yields a set of paraphrases, denoted by $p_1^c$, $\ldots$, $p_n^c$. Next, we make pair-wise comparisons of these paraphrases with $c$, resulting in $c-p_1^c$, $\ldots$, and $c-p_n^c$. At this step, we identify the examples which are entailed, and only those are chosen. 

However, there are many other secondary factors, for e.g., a model may only be able to generate a limited number of paraphrase variations compared to others but others can be more correct and/or consistent. As such, we considered three major dimensions in our evaluation: \textit{(i) coverage, (ii) correctness, and (iii) diversity}. To offer transparency around our experiment process, we detail the aforementioned evaluation dimensions as follows.

\begin{enumerate}
     \item \textbf{Coverage - the number of considerable paraphrase generations that a model generates:} We intend to generate up to 5 paraphrases per given claim. Given all the generated claims, we perform a minimum edit distance (MED) calculation at the word level instead of a character level. If MED is greater than $2$ for any given paraphrase candidate (for e.g., $c-p_1^c$ in the above example) with the claim then we further consider that paraphrase, otherwise discarded. We evaluated all four models based on this setup to identify the model of choice which is generating the maximum number of considerable paraphrases.
     \item \textbf{Correctness - correctness in paraphrase generations:} After the first level of filtration, we performed pairwise entailment and kept only those paraphrase candidates, marked as entailed by the \cite{}, SoTA model trained on SNLI. 
     \item \textbf{Diversity - linguistic diversity in paraphrase generations:} We are interested in choosing a model that can produce paraphrases with significant linguistic diversity. This implies that we are interested in checking for dissimilarities between generated paraphrase claims. For e.g., $p_1^c-p_2^c$, $p_1^c-p_3^c$, $p_1^c-p_4^c$, $\ldots$, $p_1^c-p_n^c$ -- this process is repeated for all the other paraphrases and the dissimilarity score is averaged across all paraphrase generations. Since there is no standard metric to measure dissimilarity, we use the inverse of the BLEU score as a proxy metric. This gives us an understanding of the linguistic diversity of a given model.
 \end{enumerate}

Based on our experiments centred around the above dimensions, we experimented with three models: (i) GPT3-text-davinci-003 \cite{brown2020language}, (ii) Pegasus \cite{zhang2020pegasus}, and (ii) ChatGPT \cite{schulman2022chatgpt} and found that GPT3-text-davinci-003 was ideal. The results of our experiments are reported in table \ref{tab:para-eval} below.

\begin{minipage}{0.45\textwidth}
\begin{table}[H]
\centering
\resizebox{\columnwidth}{!}{%
\begin{tabular}{@{}lccc@{}}
\toprule
Model           &  Coverage  & Correctness & Diversity \\ \midrule
Pegasus          &   32.46   &    94.38\%       &     3.76      \\
T5               &  30.26       &      83.84\%       &    3.17       \\
GPT3-text-davinci-003 &   35.51   &   88.16\%      &     7.72      \\
\bottomrule
\end{tabular}%
}
\caption{Evaluation dimensions of textual claim paraphrasers.}
\label{tab:para-eval}
\end{table}  
\end{minipage} \hfill
\begin{minipage}{0.45\textwidth}
    \begin{figure}[H]
    \centering
    \includegraphics[width=\columnwidth,  trim = {0 0.5cm 0 0}]{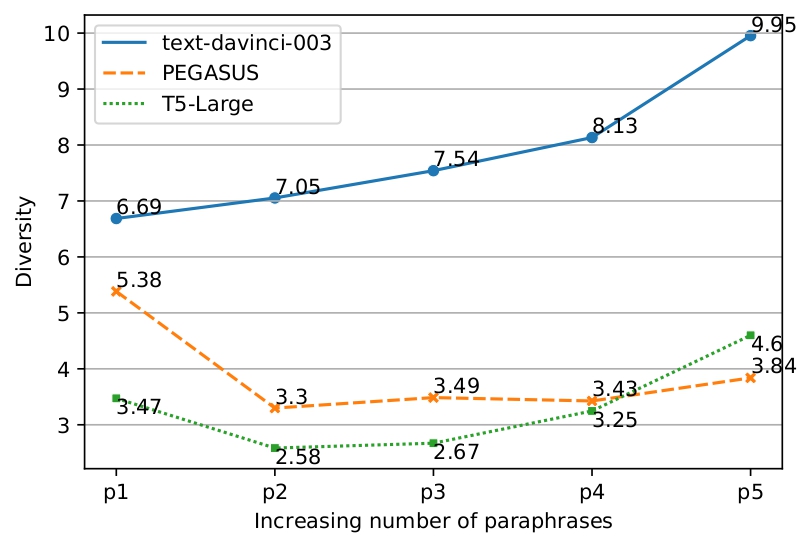}
    \caption{Variation of Diversity with Increase in number of paraphrases generated}
    \label{fig: parr}
    \end{figure}
\end{minipage}

\section{Visual paraphrasing using Stable Diffusion}\label{sec:app-B}

Building upon section \ref{sec:stable-diffusion}, we highlight the process behind visual paraphrasing in this section. Diffusion models are machine learning models that are trained to denoise random gaussian noise step by step to get a sample of interest, such as an image. However one of the major downsides of diffusion models is that the denoising process is both time and memory consumption are very expensive. The main reason for this is that they operate in pixel space which becomes unreasonably expensive, especially when generating high-resolution images. Stable diffusion was introduced to solve this problem as it depends on Latent diffusion. Latent diffusion reduces the memory and computational cost by applying the diffusion process over a lower dimensional latent space instead of on the actual pixel space. It is trained with the objective of “removing successive applications of Gaussian noise to training images”, and can be considered as a sequence of denoising autoencoders.

Quality control is a big reason to worry when paraphrasing automatically. There are two aspects we have tested for the available models - (i) variations, and (ii) the number of paraphrases generated.

\subsection{Explainability of generated images}

\section{Assessment of Stable Diffusion generated images}
While Stable Diffusion has received great acclaim owing to its stellar performance for a variety of use cases, to our knowledge, we are the first to adopt it for fake news generation. As such, to assess the quality of generated images in the context of the fake news generation task, we utilize two evaluation metrics. 

\subsection{FID \& Relevance Score-based quantitative assessment of Stable Diffusion generated images}\label{quant_measures}

While Stable Diffusion has received great acclaim owing to its stellar performance for a variety of use cases, to our knowledge, we are the first to adopt it for fake news generation. As such, to assess the quality of generated images in the context of the fake news generation task, we utilize two evaluation metrics - i) FID \cite{fid} and ii) Relevance Score \cite{hao2022optimizing} - details are discussed in the following paragraphs.

\textbf{FID Score:} In order to compute the FID scores, we first filter out the claims from our dataset that consists of person entities by leveraging the BERT-base-NER model. Following the process adopted in \cite{borji2022generated}, we ran the Mediapipe \cite{mediapipe} face detector twice: first on the entire image to detect faces, and thereafter on the individual detections to prune false positives, to extract faces from the real and Stable Diffusion generated images corresponding to the filtered set of claims. We then compute the FID between the set of faces extracted from the real and Stable Diffusion generated images using the \texttt{clean-fid} package released by \cite{clean-fid}.

\begin{figure}[!tbh]
\centering
 \begin{subfigure}{.5\textwidth}
  \centering
  \includegraphics[width=\linewidth]{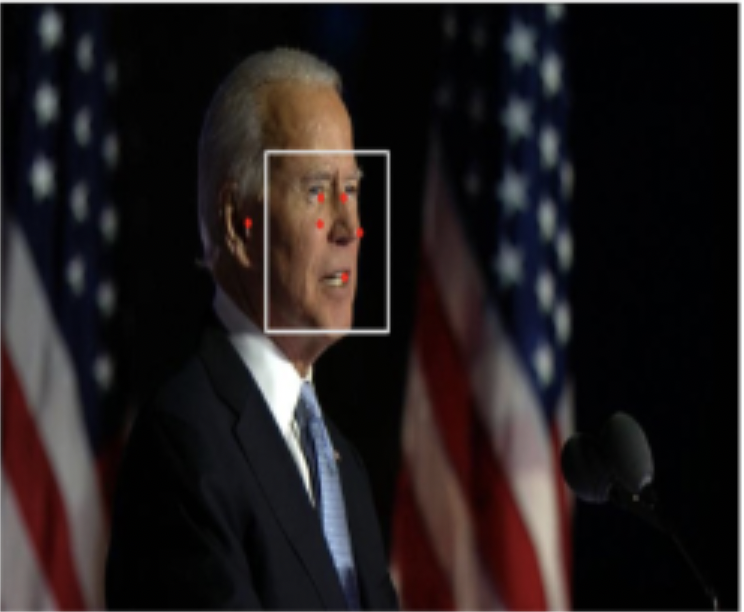}
  \caption{Real Image}
  \label{fig:sub1}
\end{subfigure}%
\begin{subfigure}{.5\textwidth}
  \centering
  \includegraphics[width=\linewidth]{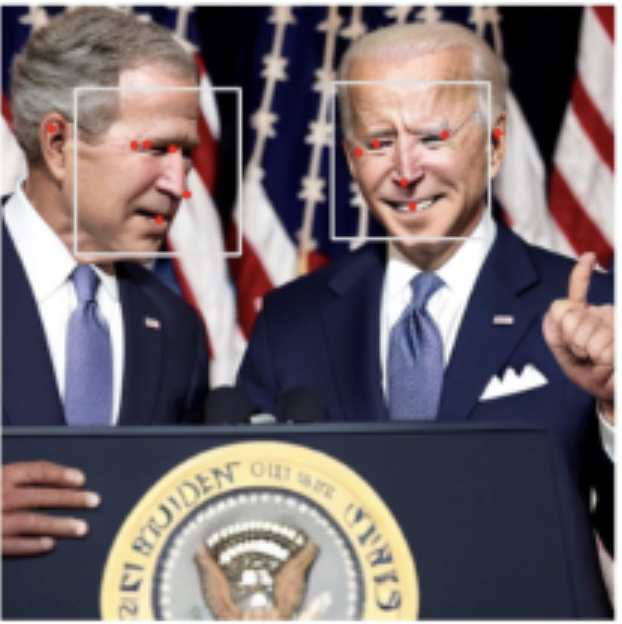}
  \caption{Stable Diffusion -v2 generated image}
  \label{fig:sub2}
\end{subfigure}
\caption{In this example, for the claim : "Former President George W. Bush congratulates President-elect Joe Biden, says election was 'fundamentally fair' and 'its outcome is clear' ", the left image where only Joe Biden is visible is the original claim image, and on the right where George Bush and Joe Biden are visible is the SD generated one. To assess the quality of the generated image we have calculated the pair-wise FID score. First, we extract the faces using Mediapipe \cite{mediapipe} Face Detector for the real and Stable Diffusion generated image for each claim. We then compute the FID using clean-fid \cite{clean-fid} pairwise. Then for a set of $500$ randomly selected samples, we average out the pairwise FID scores. It is 8.67, demonstrating a good match overall. The difference between the left image vs. the right one is the number of faces. In such a case, we take the best (lowest) FID score as the FID score for that claim. In this way, we make sure what is the common minimum between an AI-generated image vs. an actual news image.}
\label{fig: nocorr}
\end{figure}

\textbf{Relevance Score: } 
Considering Figure 14.a is the original image and Figure 14.b is the SD generated image, we compute the relevance scores as the combination of two metrics - i) CLIP score \cite{CLIP_citation} (which measures how relevant the generated image is to the user input prompt) - this is to measure the semantic similarity between text and image modality using pre-trained vision-language model CLIP, and ii) Aesthetic score \cite{hao2022optimizing}, obtained by employing a linear estimator on top of a frozen CLIP model, that is trained by human ratings in the Aesthetic Visual Analysis [MMP12] dataset. This score represents the quality of a generated image based on human evaluation pre-scores, representing what human perceives as aesthetically pleasing.

We use the relevance score as introduced in \cite{hao2022optimizing} to measure whether the generated images are relevant to the original input prompt. We compute CLIP  \cite{CLIP_citation} similarity scores to measure how relevant the generated images and the original input prompts are. The resulting relevance score is defined as:

\begin{equation}
f_{\text {rel }}(\boldsymbol{x}, \boldsymbol{y})=\mathbb{E}_{i_{\boldsymbol{y}} \sim \mathcal{G}(\boldsymbol{y})}\left[\min \left(20 * g_{\text {CLIP }}\left(\boldsymbol{x}, i_{\boldsymbol{y}}\right)-5.6,0\right)\right]
\end{equation}

where, $i_{\boldsymbol{y}} \sim \mathcal{G}(\boldsymbol{y})$ means sampling images $i_{\boldsymbol{y}}$ from the text-to-image model $\mathcal{G}$ with $\boldsymbol{y}$ as input prompt, and $g_{\text {CLIP }}(\cdot, \cdot)$ stands for the CLIP similarity function. 

Second, we employ aesthetic predictor as discussed in \cite{hao2022optimizing} to quantify aesthetic preferences. The aesthetic predictor \cite{schuhmann_2022} builds a linear estimator on top of a frozen CLIP model, which is trained by human ratings in the Aesthetic Visual Analysis [MMP12] dataset. The aesthetic score is defined as:

\begin{equation}
f_{\text {aes }}(\boldsymbol{x}, \boldsymbol{y})=\mathbb{E}_{i_{\boldsymbol{x}} \sim \mathcal{G}(\boldsymbol{x}), i_{\boldsymbol{y}} \sim \mathcal{G}(\boldsymbol{y})}\left[g_{\text {aes }}\left(i_{\boldsymbol{y}}\right)-g_{\text {aes }}\left(i_{\boldsymbol{x}}\right)\right]
\end{equation}

where, $g_{\text {aes }}(\cdot)$ denotes the aesthetic predictor, and $i_{\boldsymbol{y}}, i_{\boldsymbol{x}}$ are the images generated by the prompts $\boldsymbol{y}$ and $\boldsymbol{x}$, respectively. 
\subsection{MOS-based quality assessment of Stable Diffusion generated images}\label{apx-mos}

This section delineates the process followed to assess the quality of synthetically generated images, given the prompt used for a generation as context to the human rater. Specifically, we asked $10$ raters to assign an integral score from 1 (\textit{bad quality}) to 5 (\textit{excellent quality}) to the generated images in the context of the given prompt. Specifically, similar to \cite{chambon2022adapting}, the scoring system was verbalized as follows: 
\begin{etaremune}
    \item Life-like generated image with potentially minor error elements, but practically indistinguishable from an original.
    \item Good generated image with noticeable errors not influencing the claim's veracity assessment.
    \item Moderate errors in the generated image with possible minor negative imacpts to the claim's veracity assessment.
    \item Errors leading to hallucinated lesions while still preserving the major theme of the claim but influencing the claim's veracity assessment.
    \item Severe errors such as the generated image not following the prompt's major theme resulting in the claim's veracity assessment impossible.
\end{etaremune}

The raters rated the CLIP re-ranked output for each prompt (so $500$ images in total), presented in a randomized fashion. As part of a pilot study, we assessed the calibration procedure and the test-retest reliability of $10$ raters on a subset of $500$ generated images by adding a generated image twice to a larger test set, similar to \cite{ledig2017photo}. We observed good reliability and no significant differences between the ratings of the identical images.

\section{5W SRL}\label{sec:app-5W}
 A typical SRL system first identifies verbs in a given sentence  and then marks all the related words/phrases haven relational projection with the verb and assigns appropriate roles. Thematic roles are generally marked by standard roles defined by the Proposition Bank (generally referred to as PropBank) \cite{palmer2005proposition}, such as: \textit{Arg0, Arg1, Arg2}, and so on. We propose a mapping mechanism to map these PropBank arguments to 5W semantic roles (refer to the conversion table \ref{tab:deppar-SRL}). 

Not necessarily all the Ws are present in all the sentences. To understand this sparseness, a detailed analysis of the presence of each of the 5W at the sentence level has been done and reported in figure \ref{fig: nocorr}.

\begin{minipage}{0.5\textwidth}
    \begin{table}[H]
    \centering
    \resizebox{0.8\textwidth}{!}{
    \begin{tabular}{cccccc}
    \toprule 
    \textbf{PropBank Role }& \textbf{Who} & \textbf{What} & \textbf{When} & \textbf{Where} & \textbf{Why} \\
    \midrule 
    ARG0 & \textbf{84.48} & 0.00 & 3.33 & 0.00 & 0.00 \\
    ARG1 & 10.34 & \textbf{53.85} & 0.00 & 0.00 & 0.00 \\
    ARG2 & 0.00 & 9.89 & 0.00 & 0.00 & 0.00 \\
    ARG3 & 0.00 & 0.00 & 0.00 & 22.86 & 0.00 \\
    ARG4 & 0.00 & 3.29 & 0.00 & 34.29 & 0.00 \\
    ARGM-TMP & 0.00 & 1.09 & \textbf{60.00} & 0.00 & 0.00 \\
    ARGM-LOC & 0.00 & 1.09 & 10.00 & \textbf{25.71} & 0.00 \\
    ARGM-CAU & 0.00 & 0.00 & 0.00 & 0.00 & \textbf{100.00} \\
    ARGM-ADV & 0.00 & 4.39 & 20.00 & 0.00 & 0.00 \\
    ARGM-MNR & 0.00 & 3.85 & 0.00 & 8.57 & 0.00 \\
    ARGM-MOD & 0.00 & 4.39 & 0.00 & 0.00 & 0.00 \\
    ARGM-DIR & 0.00 & 0.01 & 0.00 & 5.71 & 0.00 \\
    ARGM-DIS & 0.00 & 1.65 & 0.00 & 0.00 & 0.00 \\
    ARGM-NEG & 0.00 & 1.09 & 0.00 & 0.00 & 0.00 \\
    \bottomrule
    \end{tabular}
    }
    \caption{A mapping table from PropBank\cite{palmer2005proposition} {(\textit{Arg0, Arg1, ...})} to 5W {(\textit{who, what, when, where, and why})}.}
    \label{tab:deppar-SRL}
    \end{table}
\end{minipage}
\begin{minipage}{0.5\textwidth}   
    \begin{figure}[H]
    \centering
    \includegraphics[width=0.9\columnwidth, trim = {0 1.5cm 0 0}]{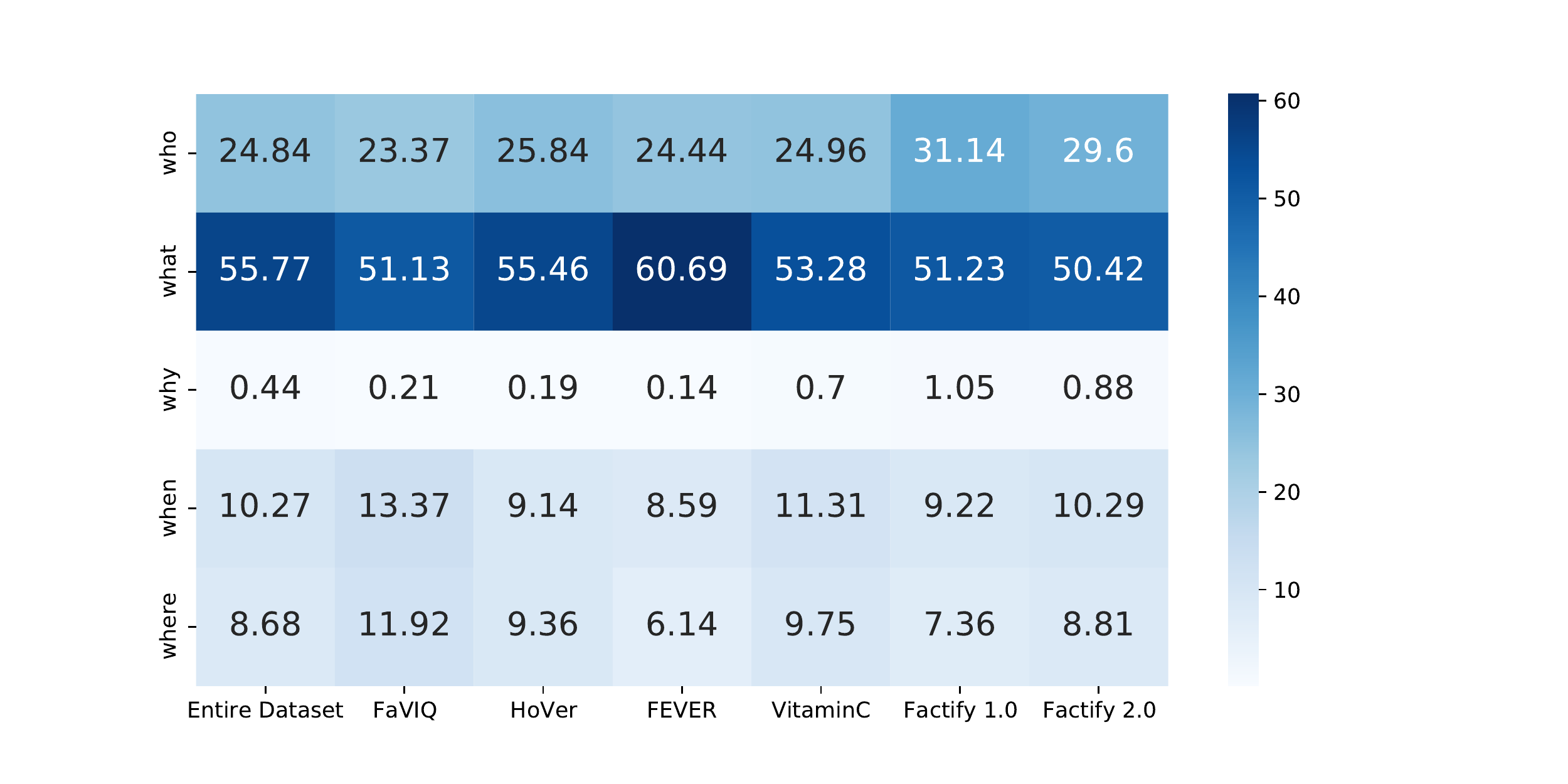}
    \caption{Sentence level co-occurrence of Ws at sentence level across the corpus.}
    \label{fig: nocorr}
    \end{figure}
\end{minipage}

\subsection{Human evaluation of 5W SRL}
In this study evaluation for the 5W Aspect, based on semantic role labelling is conducted using \textit{mapping accuracy}. This involves accuracy on SRL output mapped with 5Ws.

For the purpose of finding how good the mapping of 5W with semantic roles and generation of semantic roles, human annotation of $3500$ data points was conducted, $500$ random datapoints from the entire dataset, 500 each from FEVER \cite{thorne2018fever}, FavIQ \cite{park2021faviq}, HoVer \cite{jiang2020hover}, ViTC \cite{schuster2021get}, Factify 1.0 \cite{mishra2022factify} and Factify 2.0 \cite{surya2023factify2}, see table ~\ref{tab:srl-human eval}

\section{5W QA pairs generation using language model}\label{subsec:app-CA}
For the QG task, we shortlisted two pre-trained top-performing models for question generation according to the papers with code leaderboard where the model and code have been released.  These models were fine-tuned on various SQuAD datasets \cite{rajpurkar2018know} by simply appending the answer to the context. A random sampling on 352k data points was done to get of 15\% of the datapoint to find the best question-generating model with respect to 5W. For example, given an answer from "who" based on semantic role labeller and context from the claim, it should generate questions containing "who" and not other Ws. By modelling the claims as context and the outputs from the SRL models as answers, the process of generating 5W questions for the task of fact verification was accomplished. The pre-trained models we utilized for QG are as follows: 

\begin{itemize}
\item \textbf{BART:} BART \cite{lewis2019bart} is a denoising autoencoder for pretraining sequence-to-sequence models, trained by (i) corrupting text with an arbitrary noising function, and (ii) learning a model to reconstruct the original text. BART was trained to generate questions in two ways: casual generation and context-based generation. For this task, we used the \texttt{bart-squad-qg-hl} variant focusing on context-based generation. This variant of BART scored 24.15, 25.43, and 52.64 on the BLEU4 \cite{papineni2002bleu}, METEOR \cite{banerjee2005meteor}, and ROUGE-L metrics \cite{rouge}, respectively, whereas the current state-of-the-art (SoTA) of the BART model from Textbook 2.0 scores $25.08$, $26.73$, and $52.55$ on the same metrics. 
\item \textbf{ProphetNet:} ProphetNet \cite{qi2020prophetnet} is a generative model that uses multi-lingual pre-training with masked span generation to create shared latent representations across languages. It generates all the masked spans together, given an input sequence, and uses a future n-gram loss to prevent overfitting on strong local correlations. ProphetNet is optimized through an $n$-step look-ahead prediction, which predicts the next $n$ tokens based on previous context tokens at each time step, encouraging the model to explicitly plan for future tokens. It was evaluated on benchmarks for abstractive summarization and question generation tasks such as CNN/DailyMail, Gigaword, and SQuAD 1.1 \cite{rajpurkar2016squad}. ProphetNet has a 12-layer encoder and 12-layer decoder with $1024$ embedding/hidden size and 4096 feed-forward filter size. The batch size and training steps were set to 1024 and 500K, respectively, and Adam optimization was used with a learning rate of 3 $\times$ 10$^{-4}$ for pre-training. The input length was set to 512 and masking was done randomly in continuous spans every 64 tokens, with 15\% of the total number of tokens masked.
\end{itemize}

5W QA pair generation is a result of two submodules: (i) 5W SRL, and (ii) 5W-based QA pair generations. We have used pretrained models of context-based question generation models, wrapped in automation infrastructure. Contexts are the actual claim, and the answers are the Semantic Role Labeling outputs. As an  example, let's consider a claim, ``After April 11, 2020, there was a fatality rate of over 1.61 in Malaysia during the coronavirus pandemic". After applying SRL, we obtain the answer to the ``When" of the input sentence, yielding ``After April 11, 2020". Next, we feed the answer obtained in the prior step (After April 11, 2020) along with the context (``After April 11, 2020, there was a fatality rate of over 1.61 in Malaysia during the coronavirus pandemic.") as the input to the model. Finally, this yields a question starting with "When", which in this case is "When was the COVID-19 pandemic?".

\subsection{Human evaluation of 5W SRL and QA generation}
For the evaluation purpose, a random sample of $3500$ data points was selected for annotation. The questions generated using the Prophetnet model were utilized for this purpose. The annotators were instructed to evaluate the question-answer pairs in three dimensions: the question is well formed, which means it is syntactically correct, the question is correct which  means it is semantically correct with respect to the given claim, and extracted answer from the model is correct. The evaluation results for the datasets are presented in the following analysis, see table ~\ref{tab:human-QAG}

\begin{table*}[!h]
    
    \begin{minipage}{.5\linewidth}
      \centering
        \resizebox{\columnwidth}{!}{
      
        \begin{tabular}{cllllll}
\toprule
\multicolumn{1}{l}{} & FaVIQ               & FEVER                & HoVer                & VitaminC            & Factify1.0            & Factify2.0            \\
\hline
Who         & \cellcolor[HTML]{E4DC4D}89\% & \cellcolor[HTML]{FFDE50}85\%  & \cellcolor[HTML]{D6D94C}90\%  & \cellcolor[HTML]{FFE24F}87\% & \cellcolor[HTML]{FFE050}86\% & \cellcolor[HTML]{FFD950}82\% \\
What        & \cellcolor[HTML]{FFDE50}85\% & \cellcolor[HTML]{FFAF51}56\%  & \cellcolor[HTML]{FFC250}68\%  & \cellcolor[HTML]{FFD350}78\% & \cellcolor[HTML]{FFD850}81\% & \cellcolor[HTML]{ACCF49}93\% \\
When        & \cellcolor[HTML]{FFE050}86\% & \cellcolor[HTML]{D6D94C}90\%  & \cellcolor[HTML]{90C948}95\%  & \cellcolor[HTML]{66BF45}98\% & \cellcolor[HTML]{FFDB50}83\% & \cellcolor[HTML]{FFCE50}75\% \\
Where       & \cellcolor[HTML]{ACCF49}93\% & \cellcolor[HTML]{4AB842}100\% & \cellcolor[HTML]{D6D94C}90\%  & \cellcolor[HTML]{74C245}97\% & \cellcolor[HTML]{ACCF49}93\% & \cellcolor[HTML]{FFE050}86\% \\
Why         & \cellcolor[HTML]{FF5353}0\%  & -                             & \cellcolor[HTML]{4AB842}100\% & \cellcolor[HTML]{BAD24A}92\% & \cellcolor[HTML]{FFE24F}87\% & \cellcolor[HTML]{ACCF49}93\%

\\
\bottomrule
\end{tabular}
}
\caption{Human evaluation of 5W SRL; It is observed that for most of the datapoints \emph{why} is missing}
\label{tab:srl-human eval}
    \end{minipage}%
    \begin{minipage}{.5\linewidth}
      \centering
\resizebox{\columnwidth}{!}{

        \begin{tabular}{ccllllll}
\toprule
\multicolumn{1}{l}{}    & \multicolumn{1}{l}{}    & \multicolumn{1}{c}{FaVIQ}    & \multicolumn{1}{c}{FEVER}    & \multicolumn{1}{c}{HoVer}     & \multicolumn{1}{c}{VitaminC} & \multicolumn{1}{c}{Factify 1.0} & \multicolumn{1}{c}{Factify 2.0} \\
\hline
                        & Question is well-formed & \cellcolor[HTML]{FFE24F}86\% & \cellcolor[HTML]{FFD350}77\% & \cellcolor[HTML]{FFDE50}84\%  & \cellcolor[HTML]{FFD650}79\% & \cellcolor[HTML]{FFD850}80\% & \cellcolor[HTML]{FFDB50}82\% \\
                        & Question is correct     & \cellcolor[HTML]{D0D94F}90\% & \cellcolor[HTML]{FFDB50}82\% & \cellcolor[HTML]{FFE24F}86\%  & \cellcolor[HTML]{FFDD50}83\% & \cellcolor[HTML]{F4E04F}87\% & \cellcolor[HTML]{DCDB4F}89\% \\
\multirow{-3}{*}{Who}   & Answer is correct       & \cellcolor[HTML]{DCDB4F}89\% & \cellcolor[HTML]{FFE050}85\% & \cellcolor[HTML]{D0D94F}90\%  & \cellcolor[HTML]{F4E04F}87\% & \cellcolor[HTML]{FFE24F}86\% & \cellcolor[HTML]{FFDB50}82\% \\
\hline
                        & Question is well-formed & \cellcolor[HTML]{FFC950}71\% & \cellcolor[HTML]{FFAB51}53\% & \cellcolor[HTML]{FFC450}68\%  & \cellcolor[HTML]{FFD650}79\% & \cellcolor[HTML]{FFD350}77\% & \cellcolor[HTML]{FFCA50}72\% \\
                        & Question is correct     & \cellcolor[HTML]{FFD350}77\% & \cellcolor[HTML]{FFC550}69\% & \cellcolor[HTML]{FFC750}70\%  & \cellcolor[HTML]{FFD950}81\% & \cellcolor[HTML]{FFD850}80\% & \cellcolor[HTML]{FFD150}76\% \\
\multirow{-3}{*}{What}  & Answer is correct       & \cellcolor[HTML]{FFE050}85\% & \cellcolor[HTML]{FFB051}56\% & \cellcolor[HTML]{FFC450}68\%  & \cellcolor[HTML]{FFD450}78\% & \cellcolor[HTML]{FFD950}81\% & \cellcolor[HTML]{ACD14F}93\% \\
\hline
                        & Question is well-formed & \cellcolor[HTML]{E8DE4F}88\% & \cellcolor[HTML]{FFD350}77\% & \cellcolor[HTML]{FFE24F}86\%  & \cellcolor[HTML]{FFD450}78\% & \cellcolor[HTML]{FFD950}81\% & \cellcolor[HTML]{FFD450}78\% \\
                        & Question is correct     & \cellcolor[HTML]{D0D94F}90\% & \cellcolor[HTML]{FFE24F}86\% & \cellcolor[HTML]{E8DE4F}88\%  & \cellcolor[HTML]{A0CF4F}94\% & \cellcolor[HTML]{B8D44F}92\% & \cellcolor[HTML]{DCDB4F}89\% \\
\multirow{-3}{*}{When}  & Answer is correct       & \cellcolor[HTML]{FFE24F}86\% & \cellcolor[HTML]{D0D94F}90\% & \cellcolor[HTML]{94CD4F}95\%  & \cellcolor[HTML]{70C54F}98\% & \cellcolor[HTML]{FFDD50}83\% & \cellcolor[HTML]{FFCF50}75\% \\
\hline
                        & Question is well-formed & \cellcolor[HTML]{D0D94F}90\% & \cellcolor[HTML]{94CD4F}95\% & \cellcolor[HTML]{FFC450}68\%  & \cellcolor[HTML]{F4E04F}87\% & \cellcolor[HTML]{C4D64F}91\% & \cellcolor[HTML]{E8DE4F}88\% \\
                        & Question is correct     & \cellcolor[HTML]{FFE050}85\% & \cellcolor[HTML]{94CD4F}95\% & \cellcolor[HTML]{FFD450}78\%  & \cellcolor[HTML]{B8D44F}92\% & \cellcolor[HTML]{B8D44F}92\% & \cellcolor[HTML]{FFDD50}83\% \\
\multirow{-3}{*}{Where} & Answer is correct       & \cellcolor[HTML]{ACD14F}93\% & \cellcolor[HTML]{7CC84F}97\% & \cellcolor[HTML]{D0D94F}90\%  & \cellcolor[HTML]{7CC84F}97\% & \cellcolor[HTML]{ACD14F}93\% & \cellcolor[HTML]{FFE24F}86\% \\
\hline
                        & Question is well-formed & \cellcolor[HTML]{FF5353}0\%  & -                            & \cellcolor[HTML]{58C050}100\% & \cellcolor[HTML]{B8D44F}92\% & \cellcolor[HTML]{B8D44F}92\% & \cellcolor[HTML]{D0D94F}90\% \\
                        & Question is correct     & \cellcolor[HTML]{FF5353}0\%  & -                            & \cellcolor[HTML]{58C050}100\% & \cellcolor[HTML]{94CD4F}95\% & \cellcolor[HTML]{94CD4F}95\% & \cellcolor[HTML]{A0CF4F}94\% \\
\multirow{-3}{*}{Why}   & Answer is correct       & \cellcolor[HTML]{FF5353}0\%  & -                            & \cellcolor[HTML]{58C050}100\% & \cellcolor[HTML]{88CA4F}96\% & \cellcolor[HTML]{F4E04F}87\% & \cellcolor[HTML]{ACD14F}93\%

\\
\bottomrule
\end{tabular}
}

\caption{Human evaluation of QA generation}
\label{tab:human-QAG}

    \end{minipage} 
\end{table*}

\section{5W QA-based validation}
To design the 5W QA validation system, we utilized the claims, evidence documents, and 5W questions generated by the question generation system as input. The answer generated by the 5W QG model is treated as the gold standard for comparison between claim and evidence. We experimented with three models, T5-3B \cite{raffel2020exploring}, T5-Large \cite{raffel2020exploring}, and Bert-Large \cite{devlin2018bert}. The T5 is an encoder-decoder-based language model, that treats this task as text-to-text conversion, with multiple input sequences and produces an output as text. The model is pre-trained using the C4 corpus \cite{raffel2020exploring} and fine-tuned on a variety of tasks. T5-Large employs the same encoder-decoder architecture as T5-3B \cite{raffel2020exploring}, but with a reduced number of parameters. The final model that we experimented with is the Bert-Large \cite{devlin2018bert} model, which utilizes masked language models for pre-training, enabling it to handle various downstream tasks and represent both single and pairs of sentences in a single token sequence. It is trained using MLM and a binarized next-sentence prediction task to understand sentence relationships.

\section{Selecting the best combination - 5W QAG vs. 5W QA validation}\label{sec:app-5W valn}
We have utilized off-the-self models both for 5W question-answer generation and 5W question-answer validation. Given that the datasets using for training the models bear an obvious discrepancy in terms of the distribution characteristics compared to our data (world news) which would probably lead to a generalization gap, it was essential to experimentally judge which system offered the best performance for our use-case. Instead of choosing the best system for generation vs. validation, we opted for pair-wise validation to ensure we chose the best combination. Table \ref{tab:QAG-QA} details our evaluation results -- the rows denote the QA models while the columns denote QAG models. From the results in the table, we can see that the best combination in terms of a QAG and QA validation model was identified as T5-3b and ProphetNet respectively.

\begin{table}[H]
\centering
\resizebox{\columnwidth}{!}{%
\begin{tabular}{@{}lllllllllllllllll@{}}
\toprule
           & \multicolumn{8}{c}{ProphetNet}                                  & \multicolumn{8}{c}{BART}                                        \\ \midrule
           & \multicolumn{4}{c}{Claim}     & \multicolumn{4}{c}{+Paraphrase} & \multicolumn{4}{c}{Claim}     & \multicolumn{4}{c}{+Paraphrase} \\
 &
  \multicolumn{1}{c}{BLEU} &
  \multicolumn{1}{c}{ROUGHEL} &
  \multicolumn{1}{c}{Recall} &
  \multicolumn{1}{c}{F1} &
  \multicolumn{1}{c}{BLEU} &
  \multicolumn{1}{c}{ROUGHEL} &
  \multicolumn{1}{c}{Recall} &
  \multicolumn{1}{c}{F1} &
  \multicolumn{1}{c}{BLEU} &
  \multicolumn{1}{c}{ROUGHEL} &
  \multicolumn{1}{c}{Recall} &
  \multicolumn{1}{c}{F1} &
  \multicolumn{1}{c}{BLEU} &
  \multicolumn{1}{c}{ROUGHEL} &
  \multicolumn{1}{c}{Recall} &
  \multicolumn{1}{c}{F1} \\
T5-3b &
  \cellcolor[HTML]{9AFF99}\textbf{29.22} &
  \cellcolor[HTML]{9AFF99}\textbf{48.13} &
  \cellcolor[HTML]{9AFF99}\textbf{35.66} &
  \cellcolor[HTML]{9AFF99}\textbf{38.03} &
  \cellcolor[HTML]{9AFF99}\textbf{28.13} &
  \cellcolor[HTML]{9AFF99}\textbf{46.18} &
  \cellcolor[HTML]{9AFF99}\textbf{34.15} &
  \cellcolor[HTML]{9AFF99}\textbf{36.62} &
  21.78 &
  34.53 &
  28.03 &
  28.07 &
  20.93 &
  33.57 &
  27.65 &
  27.24 \\
T5-Large   & 28.81 & 48.02 & 35.26 & 37.81 & 21.46  & 46.45  & 27.19 & 36.76 & 21.46 & 34.90 & 27.41 & 27.99 & 20.88  & 33.69  & 20.88 & 27.31 \\
BERT large & 28.65 & 46.25 & 34.55 & 36.72 & 27.27  & 44.10  & 32.95 & 35    & 20.66 & 33.19 & 25.51 & 26.44 & 19.74  & 32.34  & 25.14 & 25.71 \\ \bottomrule
\end{tabular}
}
\caption{Selecting the best combination - 5W QAG vs. 5W QA validation}
\label{tab:QAG-QA}
\end{table}

\section{Injecting adversarial assertion for fake news}\label{subsec:app-adv-assertion}

The extraordinary capabilities of today's large language models to generate realistic text based on prompts has had an electrifying impact on the scientific community. Per \cite{chat-fake}, ``Human reviewers could only detect fake abstracts [of scientific articles] 68\% of the time''. Given these major advances in language models, it is even easier today to generate and propagate misinformation in the form of fake news that would be extremely difficult, even for human experts, to detect as false without the proper tools to verify its authenticity. 

We have thus included some fake news claims synthetically generated by OPT in our dataset to provide a more realistic view of news media in recent times. This adversarial attack would help build more robust fact verification models if they are able to detect these fake claims. 

\begin{minipage}{0.48\textwidth}
    \begin{figure}[H]
        \centering
        \includegraphics[width=\columnwidth,  trim = {0 1cm 0 0}]{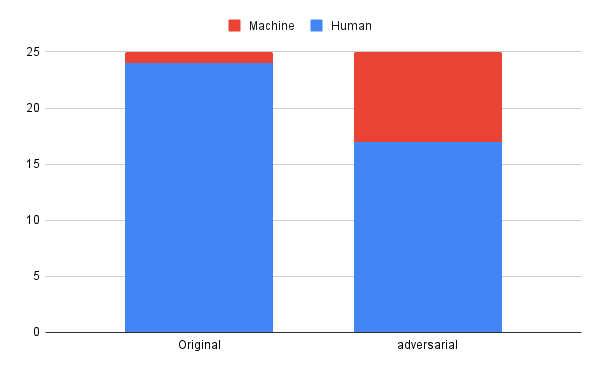}
        \caption{Representation of Human vs Machine}
        \label{fig:gptdetect1}
    \end{figure}
\end{minipage}
\begin{minipage}{0.48\textwidth}
    \begin{figure}[H]
        \centering
        \includegraphics[width=\columnwidth,  trim = {0 1cm 0 0}]{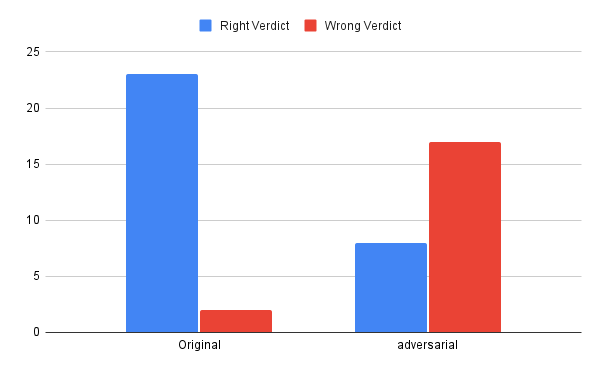}
        \caption{Representation of Right vs Wrong verdicts}
        \label{fig:gptdetect2}
    \end{figure}
\end{minipage}

\end{document}